\pdfoutput=1

\documentclass[11pt]{article}

\usepackage{acl}
\usepackage{amssymb}
\usepackage{amsmath}
\usepackage{graphicx}
\usepackage{multirow,booktabs}
\usepackage{xcolor}
\usepackage{CJKutf8}
\usepackage{amsthm}
\usepackage{mathrsfs}
\usepackage{bm}

\usepackage{times}
\usepackage{latexsym}

\usepackage[T1]{fontenc}

\usepackage[utf8]{inputenc}

\usepackage{microtype}

\usepackage{inconsolata}

\usepackage{graphicx}
\usepackage{tikz-qtree}
\usetikzlibrary{trees,decorations.text,math,calc, positioning, arrows.meta}

\title{Merge, Ensemble, and Cooperate! A Survey on Collaborative Strategies in the Era of Large Language Models}

\author{%
  Jinliang Lu$^{1,2}$$^{*}$, Ziliang Pang$^{1}$$^{*}$, Min Xiao$^{1,2}$$^{*}$, Yaochen Zhu$^{3}$ \thanks{Equal Contribution}, Rui Xia$^{3}$, Jiajun Zhang$^{1,2,4}$ \thanks{Corresponding author}  \\
  $^{1}$Institute of Automation, Chinese Academy of Sciences, Beijing, China \\
  $^{2}$School of Artificial Intelligence, University of Chinese Academy of Sciences, Beijing, China \\
  $^{3}$Nanjing University of Science and Technology, Nanjing, China \\
  $^{4}$Wuhan AI Research, Wuhan, China \\
  \texttt{\{lujinliang2019, ziliang.pang\}@ia.ac.cn}, \\ \texttt{\{yczhu, rxia\}@njust.edu.cn}, \texttt{\{jjzhang, min.xiao\}@nlpr.ia.ac.cn} \\}

\begin{document}
\maketitle
\begin{abstract}
The remarkable success of Large Language Models (LLMs) has ushered natural language processing (NLP) research into a new era. Despite their diverse capabilities, LLMs trained on different corpora exhibit varying strengths and weaknesses, leading to challenges in maximizing their overall efficiency and versatility. To address these challenges, recent studies have explored collaborative strategies for LLMs. This paper provides a comprehensive overview of this emerging research area, highlighting the motivation behind such collaborations. Specifically, we categorize collaborative strategies into three primary approaches: \textit{Merging}, \textit{Ensemble}, and \textit{Cooperation}. \textit{Merging} involves integrating multiple LLMs in the parameter space. \textit{Ensemble} combines the outputs of various LLMs. \textit{Cooperation} leverages different LLMs to allow full play to their diverse capabilities for specific tasks. We provide in-depth introductions to these methods from different perspectives and discuss their potential applications. Additionally, we outline future research directions, hoping this work will catalyze further studies on LLM collaborations and paving the way for advanced NLP applications.
\end{abstract}

\section{Introduction}
\label{sec:introduction}
\textit{"Many hands make light work."}

\begin{flushright} ~-----~ \textbf{John Heywood} \end{flushright}

\noindent 

\noindent Human beings have long understood the power of collaboration. When individuals pool their diverse skills and efforts, they can achieve far more than they could alone. This principle of collective effort has found new relevance in the realm of machine learning \cite{dietterich2000ensemble, panait2005cooperative, sagi2018ensemble}, significantly boosting the development of artificial intelligence.

\begin{figure}[t]
    \centering
    \includegraphics[scale=0.24]{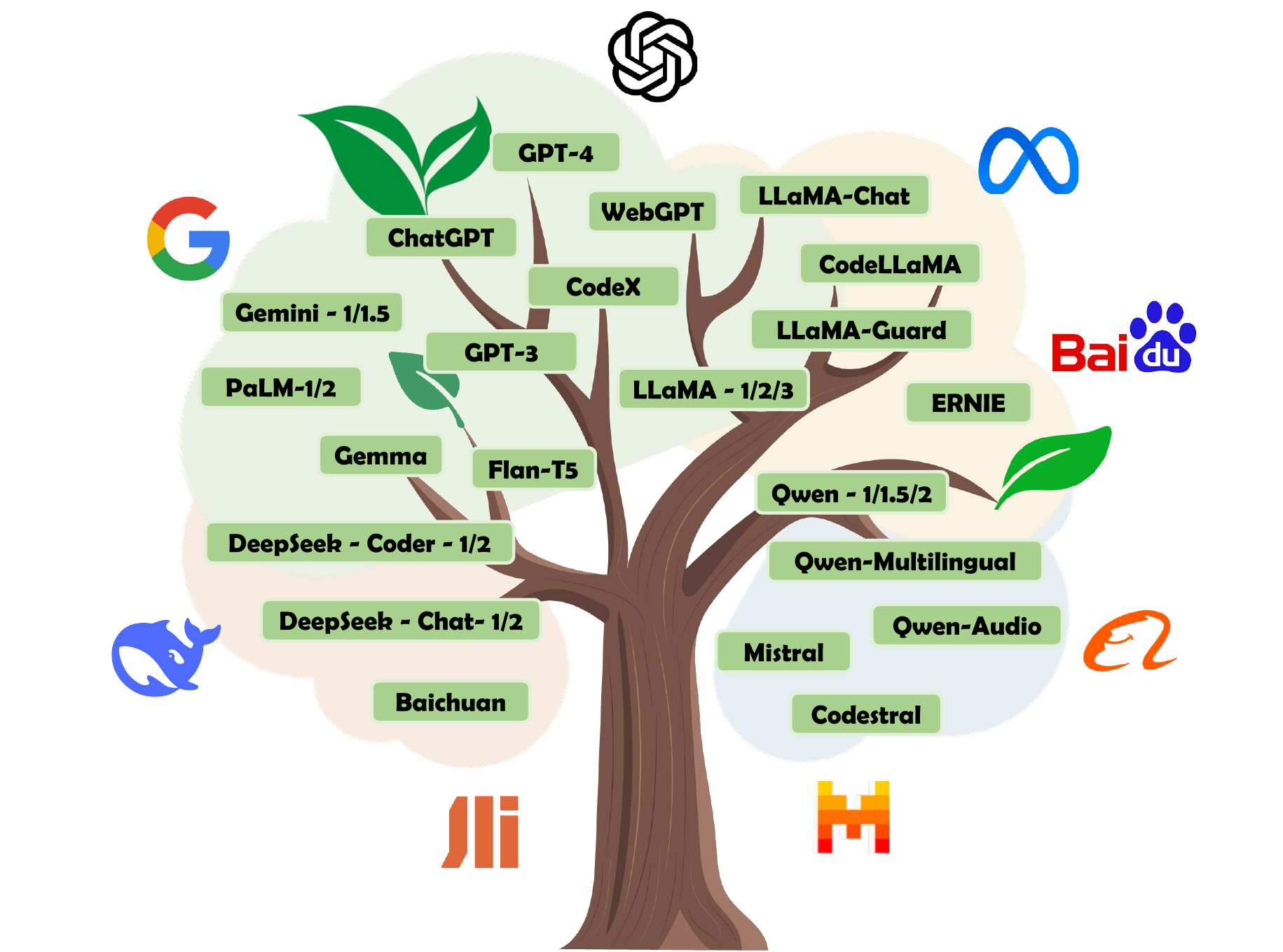}
    \caption{Recently, numerous large language models have been released, each with its own unique strengths. This diversity has fueled research into collaboration between these models.}
    \label{fig.intro_illustration}
\end{figure}

In recent years, large language models (LLMs) \cite{brown2020language, chowdhery2023palm} have emerged as one of the most rapidly developing and promising directions in artificial intelligence. These models have significantly transformed the paradigm of natural language processing (NLP) \cite{min2023recent, chang2024survey, zhao2023survey} and influenced other areas \cite{wu2023multimodal, zhang2024vision}. This impressive revolution has inspired numerous universities, institutes, and companies to pre-train and release their own LLMs. Currently, over 74,000 pre-trained models are available on the HuggingFace model hub\footnote{\url{https://huggingface.co/models}}. As shown in Figure \ref{fig.intro_illustration}, these models, trained with diverse data, architectures, and methodologies, possess unique capabilities: some are proficient in multilingual tasks \cite{le2023bloom, lin-etal-2022-shot}, others specialize in domains like medicine \cite{yang2024zhongjing} or finance \cite{wu2023bloomberggpt}, some are adept at processing long-context windows \cite{chen2023extending, chen2023longlora}, while others are fine-tuned for better alignment with human interaction \cite{ouyang2022training}. However, no single model consistently outperforms all others across tasks \cite{jiang2023llm}. This variability motivates research into the collaboration between various LLMs to unlock their combined potential, akin to creating a \textit{Hexagon Warrior}.

Despite progress in LLM collaboration research, the relationships and context among the proposed methods remain unclear. This survey aims to fill that gap by categorizing collaboration techniques into three main approaches: \textbf{\textit{Merging}}, \textbf{\textit{Ensemble}}, and \textbf{\textit{Cooperation}}. Specifically, \textbf{\textit{Merging}} and \textbf{\textit{Ensemble}} methods for LLMs are derived from traditional fusion techniques commonly explored in machine learning \cite{li2023deep}. These methods are tailored to be more suitable for LLMs, effectively leveraging the collaborative advantages of diverse LLMs.
\textbf{\textit{Merging}} involves integrating the parameters of multiple LLMs into a single, unified model, requiring that the parameters are compatible within a linear space.
In contrast, \textbf{\textit{Ensemble}} focuses on combining the outputs generated by various LLMs to produce coherent results, with less emphasis on the parameters of the individual models.
\textit{\textbf{Cooperation}} extends beyond \textit{merging} and \textit{ensemble}. This survey concentrates on cooperative methods that harness the diverse strengths of LLMs to achieve specific objectives. In general, these techniques expand the methodologies for model collaboration, holding significant research importance for LLMs.

The structure of this work is organized as follows. We begin by providing the background of LLMs and defining collaboration techniques for LLMs in Section \ref{sec:background}. Next, we introduce three key categories: \textit{\textbf{Merging}} in Section 3, \textit{\textbf{Ensemble}} in Section \ref{sec:ensemble}, and \textit{\textbf{Cooperation}} in Section \ref{sec:cooperation}. Each category of methods is thoroughly classified and described in detail, offering a clear understanding of their respective frameworks and applications. Finally, we offer a comprehensive discussion in Section \ref{sec:future}, highlighting challenges and future directions for research.

In summary, this study aims to comprehensively explore the strategies and methodologies for collaborative efforts among LLMs. We aspire for this survey to enrich understanding of LLM collaboration strategies and to inspire future research.

\begin{figure*}[h]
    \centering
    \includegraphics[width=1.0\linewidth,scale=0.80]{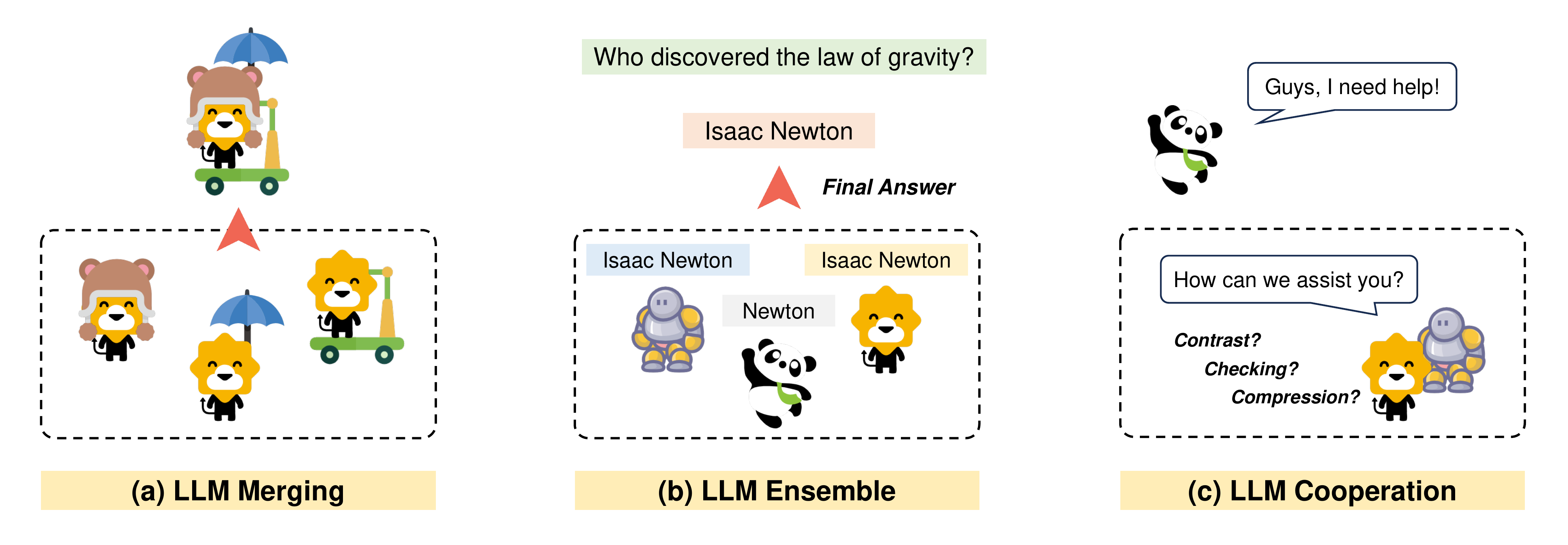}
    \caption{The illustration of different collaboration strategies, with each animal in the figures representing a different LLM.}
    \label{fig.collaboration_classification}
\end{figure*}

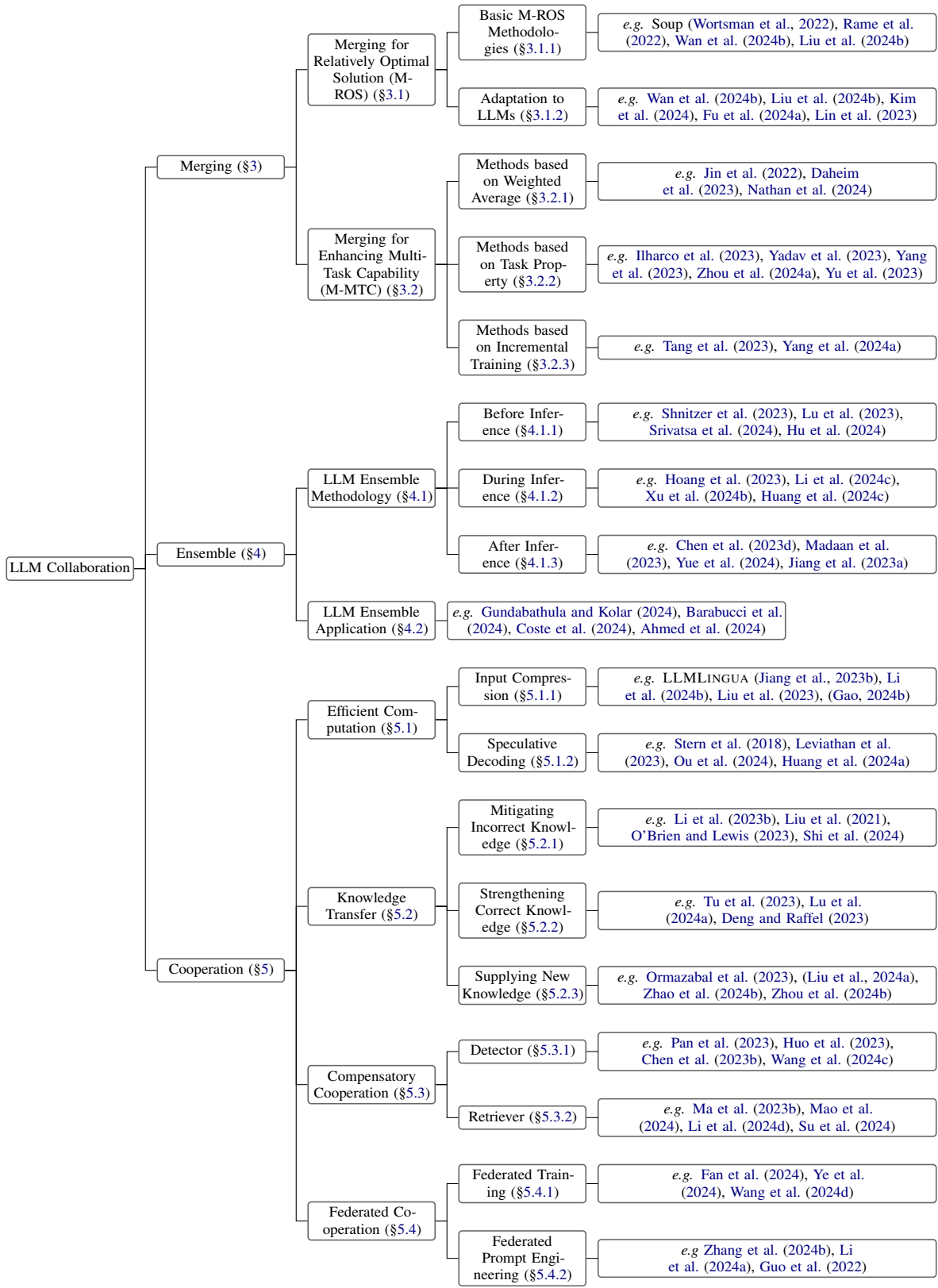
\begin {figure*}
\centering
\begin{tikzpicture}
 
\tikzset{
  grow'=right,level distance=25mm, sibling distance =4.5mm,
  execute at begin node=\strut,
  every tree node/.style={
			draw=gray!80!black,
            line width=0.6pt,
            text width=2.0cm,
			rounded corners=1.5pt,
			anchor = west,
            fill=white,
			minimum width=2mm,
			inner sep=1.5pt,
			align=center,
			font = {\scriptsize}},
         edge from parent/.style={draw=black,
         edge from parent fork right}
}

\begin{scope}[frontier/.style={sibling distance=4em,level distance = 6em}]
\Tree
[.{LLM Collaboration}
		[.{Merging (\S \ref{sec:merging})}
			[.{Merging for Relatively Optimal Solution (M-ROS) (\S \ref{sec:merging-average})}
              [.{Basic M-ROS Methodologies (\S \ref{sec:m-ros-methods})}
                \node[fill=white,text width=5.5cm](t1){\textit{e.g.} Soup \cite{wortsman2022model}, \citet{rame2022diverse}, \citet{wan2024fusechat},
                \citet{liu2024checkpoint}};
              ]
              [.{Adaptation to LLMs (\S \ref{sec:adaptation_llms})}
                \node[fill=white,text width=5.5cm](t1){\textit{e.g.} \citet{wan2024fusechat}, \citet{liu2024checkpoint}, \citet{Kim2024Prometheus2A}, \citet{splitandmerge}, \citet{Lin2023MitigatingTA}};
              ]
			]
			[.{Merging for Enhancing Multi-Task Capability (M-MTC) (\S \ref{sec:merging-multitask})}
			  [.{Methods based on Weighted Average (\S \ref{sec:weight_average_multi_task})}
                \node[fill=white,text width=5.5cm](t1){\textit{e.g.}  \citet{jin2022dataless}, \citet{daheim2023model}, \citet{nathan2024fisher}};
              ]
              [.{Methods based on Task Property (\S \ref{sec:task_property})}
                \node[fill=white,text width=5.5cm](t1){\textit{e.g.}  \citet{task_arithmetic}, \citet{ties}, \citet{AdaMerging}, \citet{metagpt}, \citet{mario}};
              ]
              [.{Methods based on Incremental Training (\S \ref{sec:incremental_Training})}
                \node[fill=white,text width=5.5cm](t1){\textit{e.g.} \citet{Concrete}, \citet{surgery}};
              ]
			]
		]
		[.{Ensemble (\S \ref{sec:ensemble})}
			[.{LLM Ensemble Methodology (\S \ref{sec:ensemble-methodology})}
              [.{Before Inference (\S \ref{sec:ensemble-methodology-before})}
                \node[fill=white,text width=5.5cm](t1){\textit{e.g.} \citet{shnitzer2023large}, \citet{lu2023routing}, \citet{srivatsa2024harnessing}, \citet{hu2024routerbench}};
              ]
              [.{During Inference (\S \ref{sec:ensemble-methodology-during})}
                \node[fill=white,text width=5.5cm](t1){\textit{e.g.} \citet{hoang2023fly}, \citet{li2024purifying}, \citet{xu2024bridging}, \citet{huang2024enabling}};
              ]
              [.{After Inference (\S \ref{sec:ensemble-methodology-after})}
                \node[fill=white,text width=5.5cm](t1){\textit{e.g.} \citet{chen2023frugalgpt}, \citet{madaan2023automix}, \citet{yue2024large}, \citet{jiang2023llm}};
              ]
			]
			[.{LLM Ensemble Application (\S \ref{sec:ensemble-application})}
                \node[fill=white,text width=5.5cm](t1){\textit{e.g.}  \citet{gundabathula2024promptmind},  \citet{barabucci2024combining}, \citet{coste2024reward}, \citet{ahmed2024scalable}};
			]
		]
		[.{Cooperation (\S \ref{sec:cooperation})}
			[.{Efficient Computation (\S \ref{sec:cooperate-accelerate})}
              [.{Input Compression (\S \ref{sec:cooperate-accelerate-predecoding})}
                \node[fill=white,text width=5.5cm](t1){\textit{e.g.} \textsc{LLMLingua} \cite{jiang-etal-2023-llmlingua}, \citet{li2024pctoolkit}, \citet{liu-etal-2023-tcra}, \cite{gao2024unifying}};
              ]
              [.{Speculative Decoding (\S \ref{sec:cooperate-accelerate-decoding})}
                \node[fill=white,text width=5.5cm](t1){\textit{e.g.} \citet{stern2018blockwise}, \citet{leviathan2023fast}, \citet{Ou2024LosslessAO}, \citet{huang2024specdec++}};
              ]
            ]
            [.{Knowledge Transfer (\S \ref{sec:cooperate-transferring})}
              [.{Mitigating Incorrect Knowledge (\S \ref{sec:cooperate-transfer-wiping_hallucination})}
                \node[fill=white,text width=5.5cm](t1){\textit{e.g.} \citet{li2023contrastive}, \citet{liu2021dexperts}, \citet{o2023contrastive}, \citet{shi2024unchosen}};
              ]
              [.{Strengthening Correct Knowledge (\S \ref{sec:cooperate-transfer-Verifying})}
                \node[fill=white,text width=5.5cm](t1){\textit{e.g.} \citet{tu2023unlocking}, \citet{lu2024diver}, \citet{deng-raffel-2023-reward}};
              ]
              [.{Supplying New Knowledge (\S \ref{sec:cooperate-transfer-Supplying})}
                \node[fill=white,text width=5.5cm](t1){\textit{e.g.} \citet{ormazabal-etal-2023-comblm}, \cite{liu2024tuning}, \citet{zhao2024weak}, \citet{zhou2024weak}};
              ]
            ]
            [.{Compensatory Cooperation (\S \ref{sec:cooperate-compensatory})}
              [.{Detector (\S \ref{sec:cooperate-compensatory-detector})}
                \node[fill=white,text width=5.5cm](t1){\textit{e.g.} \citet{Pan2023FactCheckingCC}, \citet{Huo2023RetrievingSE}, \citet{Chen2023UnderstandingRA}, \citet{Wang2024LLMsKW}};
              ]
              [.{Retriever (\S \ref{sec:cooperate-compensatory-retriever})}
                \node[fill=white,text width=5.5cm](t1){\textit{e.g.} \citet{ma-etal-2023-query}, \citet{Mao2024RaFeRF}, \citet{li2024chainofknowledge}, \citet{su2024semistructured}};
              ]
            ]
            [.{Federated Cooperation (\S \ref{sec:cooperate-federated})}
              [.{Federated Training (\S \ref{sec:cooperation-federate-training})}
                \node[fill=white,text width=5.5cm](t1){\textit{e.g.} \citet{Fan2024FedMKTFM}, \citet{Ye2024OpenFedLLMTL}, \citet{Wang2024SaveIA}};
              ]
              [.{Federated Prompt Engineering (\S \ref{sec:cooperation-federate-prompt})}
                \node[fill=white,text width=5.5cm](t1){\textit{e.g} \citet{Zhang2024CoGenesisAF}, 
                \citet{ Li2024FederatedDK}, \citet{guo2022promptfl}};
              ]
            ]
		]
]
\end{scope}

\end{tikzpicture}
\caption{The primary categorization of LLM collaboration in this survey.}
\label{fig.main_stream}
\end{figure*}

\section{Background}
\label{sec:background}
\subsection{Large Language Models}
Language modeling has always been a cornerstone of natural language processing (NLP). Recently, plenty of studies scale up of Transformer-based language models \cite{vaswani2017attention, radford2018improving} to substantial more than billions of parameters, exemplified by models like GPT-3 \cite{brown2020language}, PaLM \cite{chowdhery2023palm, anil2023palm}, LLaMA \cite{touvron2023llama, touvron2023llama2}. These models are typically considered as \textit{Large Language Models} (LLMs) due to their massive amount of parameters \cite{zhao2023survey}. This subsection discusses the architecture and scaling of LLMs, their training objectives, and the emergent abilities they exhibit.

\paragraph{Architecture and Scaling}
Similar to pre-trained language models (PLMs) \cite{radford2018improving, devlin-etal-2019-bert}, LLMs primarily adopt the Transformer architecture \cite{vaswani2017attention} as their backbone, consisting of stacked multi-head attention and feed-forward layers. Unlike PLMs, most currently released LLMs are built upon decoder-only architectures for training efficiency and few-shot capabilities. This approach also shows potential when the number of parameters increases \cite{zhang2024examining}. Recent studies have investigated the quantitative relationship between model capacity, the amount of training data, and model size, known as the scaling law \cite{kaplan2020scaling, hoffmann2022training}.

\paragraph{Training Objectives}
In the previous studies about PLMs, various language modeling tasks are proposed. For example, \textit{masked language modeling} for BERT \cite{devlin-etal-2019-bert}, \textit{De-noising language modeling} for BART \cite{lewis-etal-2020-bart} and T5 \cite{raffel2020exploring}. However, current LLMs typically utilize the standard \textit{causal language modeling} as their training objective, which aims to predict the next token based on the preceding tokens in a sequence. This training objective is well-suited for decoder-only architectures.

Beyond the pre-training objective, recent studies have aimed to model human preferences to better align LLMs with human expectations. For example, the well-known InstructGPT \cite{ouyang2022training} introduces \textit{reinforcement learning from human feedback} (RLHF), which uses preference rewards as an additional training objective. Although RLHF is effective at making LLMs more helpful to users, it inevitably incurs an alignment tax, which refers to performance degradation after RLHF. Recent research has explored various techniques to mitigate alignment tax issues \cite{Lin2023MitigatingTA, lu2024online, fu2024disperse}.

\paragraph{Emergent Abilities}
The fundamental capability of language models is text generation, where tokens are auto-regressively generated based on preceding tokens using greedy search or nucleus sampling \cite{Holtzman2020The}:
\begin{gather}
    y_{i} \sim p(y_{i} | y_{<i})
\end{gather}

Interestingly, LLMs can not only generate realistic text but also perform specific tasks when provided with task-specific prompts, without requiring fine-tuning on particular downstream tasks \cite{brown2020language}. This phenomenon is one of the most important differences between LLMs and previous PLMs. \citet{wei2022emergent} define the emergent ability as ``\textit{an ability that is not present in smaller models but is present in larger models}.'' Among these emergent abilities, \textit{in-context learning} (ICL) \cite{brown2020language, dong2022survey} and \textit{instruction following} are commonly used and significantly enhance the ability of LLMs to process various tasks.

ICL helps LLMs understand tasks by using several task examples as demonstrations. When provide these demonstrations as prompts, LLMs can automatically generate reasonable output for the given test example, which can be formalized as:
\begin{gather}
    p(\mathbf{y} | \mathbf{x}) = p(\mathbf{y} | \mathbf{x},  \text{demonstration}(\{(\mathbf{x}_{i}, \mathbf{y}_{i})\}_{i=1}^{k}))
\end{gather}

\textit{Instruction following} ability are typically emerge in LLMs that have been fine-tuned on examples formatted with instructions on multiple tasks. The generation process can be formalized as:
\begin{gather}
    p(\mathbf{y} | \mathbf{x}) = p(\mathbf{y} | \mathbf{x}, \mathcal{I} )
\end{gather}
where $\mathcal{I}$ refers to the given instruction for current example $\mathbf{x}$. The instruction tuning technique \cite{sanh2021multitask, ouyang2022training, wei2022finetuned} can enhance the generalization capabilities of LLMs, enabling them to perform well with instructions on a variety of tasks, including unseen ones \cite{thoppilan2022lamda}. 

\subsection{Collaboration for LLMs}
For previous task-dependent NLP models, collaboration strategies typically aimed to improve performance on specific tasks \cite{jia2023review}. Recently, LLMs have revolutionized NLP by demonstrating remarkable versatility across a wide range of tasks. This shift has also shifted the focus of collaboration strategies for LLMs toward enhancing versatility and achieving more general objectives. Consequently, some recently proposed collaboration strategies have become more flexible and tailored specifically for LLMs.

\paragraph{The Necessity of LLM Collaboration}
Although almost all LLMs demonstrate strong versatility across various tasks through \textit{in-context learning} and \textit{instruction following}, different LLMs still have distinct strengths and weaknesses \cite{jiang2023llm}.

Differences in training corpora and model architectures among various LLM families—such as LLaMA, GLM \cite{zeng2023glmb}, and QWen \cite{bai2023qwen}—result in significant variations in their capabilities. Even within the same family, fine-tuning on specific corpora (\textit{e.g.}, mathematics \cite{azerbayev2023llemma}, code \cite{roziere2023code}, or medical domains \cite{wu2024pmc}) can lead to noticeable performance differences. Effective collaboration among these LLMs can unlock their full potential, significantly enhancing their overall performance and versatility.

Furthermore, LLMs inevitably suffer from computational inefficiencies \cite{Zhou2024ASO}, hallucinations \cite{rawte2023survey, ji2023survey, huang2023survey}, and privacy leaks \citet{Fan2024FedMKTFM}. Recent studies explore the collaboration strategies between LLMs, which provides potential solutions to mitigate these issues and compensate for their shortcomings.

\paragraph{The Category of LLM Collaboration Methods}
Collaboration between LLMs refers to the process where \textit{multiple LLMs work together, leveraging their individual strengths and capabilities to achieve a shared objective}. In this survey, we categorize LLM collaboration methods into three aspects: \textit{merging}, \textit{ensemble} and \textit{cooperation}. As shown in Figure \ref{fig.collaboration_classification}, 
\begin{itemize}
    \item \textit{Merging} involves integrating multiple LLMs into a unified, stronger one, primarily through arithmetic operations in the model parameter space.
    \item \textit{Ensemble} combines the outputs of different models to obtain coherent results. Recent studies have proposed various ensemble methods tailored for LLMs.
    \item \textit{Cooperation} is a relatively broad concept. This survey focuses on cooperation methods that leverage the diverse capabilities of different LLMs to accomplish specific objectives, such as efficient computation or knowledge transfer.
\end{itemize}

It should be noted that as we move from \textit{merging} to \textit{ensemble} to \textit{cooperation}, the requirements for LLMs gradually relax, making the proposed methods increasingly flexible.
Specifically, \textit{merging} methods are effective only when the LLMs share a compatible parameter space, allowing seamless integration. \textit{Ensemble} methods require LLMs to have diverse yet comparable abilities; without this balance, the ensemble may be less effective. In contrast, \textit{cooperation} methods are more flexible, focusing on leveraging LLMs with various capabilities that are specially designed to achieve particular objectives.

For each category, we further classify specific methods based on their focus or stages of implementation. The comprehensive categorization is shown in Figure \ref{fig.main_stream}.

\section{Merging}
\label{sec:merging}
Single models have inherent limitations, such as potentially missing important information \cite{sagi2018ensemble}, and being prone to getting stuck in local optima or lacking multi-task capabilities. To address these limitations, researchers have explored model \textit{merging} methods, which combine multiple models in the parameter space to create a unified, stronger model. Model merging has made significant progress in recent years, with various techniques cataloged in existing surveys \cite{li2023deep}. In the era of LLMs, model merging has become an important solution for model collaboration, usually employing basic merging methods and demonstrate the effectiveness. This section focuses on the merging techniques that are proven to be effective for LLMs\footnote{Some advanced methods, such as merging after neuron alignments \text{-} like OT Fusion \cite{singh2020model}, Re-Basin techniques \cite{pena2023re, ainsworth2023git}, and REPAIR \cite{jordan2023repair} \text{-} have not been widely explored for LLMs. We leave the implementation of these techniques on LLMs for future work.}.

Current studies on model merging typically focus on two key issues: merging to approach a relatively optimal solution (M-ROS) and merging to enhance multi-task capability (M-MTC). Research on M-ROS is based on the finding that gradient-optimized solutions often converge near the boundary of a wide flat region rather than at the central point \cite{izmailov2018averaging}. Model merging offers a way to approach this relatively optimal point, thereby yielding a stronger model. M-MTC, on the other hand, aims to utilize model merging techniques to enrich a single model with capabilities across multiple tasks \cite{task_arithmetic, ties}. In the following subsection, we will introduce the techniques for each objective and their application to LLMs.

It is important to note that for both M-ROS and M-MTC, current model merging methods are applicable only to models with the same architecture and parameters within the same space. Therefore, most candidate models for merging should be trained with identical initialization. For instance, the candidate models $\mathcal{M} = \{\mathcal{M}_{1}, \mathcal{M}_{2}, \cdots, \mathcal{M}_{k}\}$ should be fine-tuned from the same pre-trained model $\mathcal{M}_{0}$. This requirement ensures compatibility and coherence among the model parameters, promoting successful merging. Unfortunately, for models with incompatible parameters, such as LLaMA and QWen, current merging techniques are ineffective.

\subsection{Merging for Relatively Optimal Solution (M-ROS)}
\label{sec:merging-average}

Machine learning models, particularly deep learning models, often fail to achieve precisely optimal solutions during training \cite{li2023deep}. Researchers have demonstrated that the local optima of modern deep neural networks are connected by simple curves, and the paths along these curves can maintain nearly constant training and test accuracy. This indicates that different local optima in the weight space are not isolated but can be connected through low-loss paths \cite{garipov2018loss}. The model obtained by the weighted averaging method can be considered a point on the low-loss path. Parameter averaging integrates the advantages of different models by averaging various local optima, reducing the bias and variance of individual models \cite{rame2022diverse}.

To improve model performance, M-ROS methods have been proposed. These methods aim to combine the parameters of multiple models, merging relatively optimized points into a better one. We categorize these methods into two groups: \textit{Simple Average} and \textit{Weighted Average}. Although initially developed for small deep models, these techniques are also effective for LLMs. We will first introduce the basic M-ROS methodologies and then discuss their application to LLMs.

\subsubsection{Basic M-ROS Methodologies}
\label{sec:m-ros-methods}
\paragraph{Simple Average}
Simple parameter averaging is a kinds of methods used to combine the parameters of multiple fine-tuned models with equal coefficients \cite{guo2023stochastic}, thus creating a stronger one \cite{singh2020model}. Empirical evidence shows that these fused models often outperform individual models in terms of accuracy, robustness and stability. For $k$ candidate models, $\mathcal{M} = \{\mathcal{M}_{1}, \mathcal{M}_{2}, \cdots, \mathcal{M}_{k}\}$, simple parameter averaging can be formalized as:
\begin{gather}
\bm{\theta}^{*} = \frac{1}{k} \sum_{i=1}^{k}  \bm{\theta}_{\mathcal{M}_{i}}
\end{gather}
where $\bm{\theta}_{\mathcal{M}_{i}}$ refers to the parameters of $i$-th model and $\bm{\theta}^{*}$ is the merged parameters.

To maximize the benefits of multiple models, Model Soup \cite{wortsman2022model} introduces Uniform Soup and Greedy Soup. Uniform Soup simply averages the model parameters. Greedy Soup adds models to the pool one at a time, ensuring each new model either improves or maintains performance on a validation set. Similarly, DiWA \cite{rame2022diverse} ranks candidate models by their performance on the validation set and adds new models only if they enhance performance. Typically, Greedy Soup and DiWA average the parameters of selected models for inference.

\paragraph{Weighted Average}
\label{sec:weight_average}
Weighted averaging allows for the assignment of different coefficients to individual models based on their significance or quality, ensuring a better merging. For trained networks with significant weight differences, the simple averaging method often performs poorly \cite{singh2020model}. Therefore, selecting appropriate merging coefficients for different models becomes a crucial factor. 
Weighted averaging can be formalized as:
\begin{gather}
\bm{\theta}^{*} = \sum_{i=1}^{k} \alpha_{i} \cdot \bm{\theta}_{\mathcal{M}_{i}}
\end{gather}
where $\alpha_{i}$ refers to the normalized coefficient for the candidate model $\mathcal{M}_{i}$.

Currently, several methods are available to find the optimal merging coefficients. For instance, Learned Soup \cite{wortsman2022model} optimizes the mixing coefficients on a validation set to minimize the loss function, thereby combining the strengths of multiple models to enhance overall performance. \citet{matena2022merging} propose to utilize \textit{Fisher Information Matrix} to measure the importance of parameters of models fine-tuned with different random seeds, and employ the importance scores as coefficients to merge them. Furthermore, \citet{Jang2024ModelSA} propose a method based on geometric relationships, interpolating fine-tuned models based on the angular divergence between parameters.

\subsubsection{Adaptation to LLMs}
\label{sec:adaptation_llms}
The above model merging techniques have been successfully adapted to LLMs for the objective of acquiring stronger LLMs and enhancing RLHF.

\paragraph{Acquiring Stronger LLMs} To obtain stronger LLMs, some studies propose novel methods tailored for LLMs. \citet{wan2024fusechat} suggest calculating the merging coefficients based on the variation ratio of parameter matrices before and after fine-tuning. Similarly, \citet{liu2024checkpoint} propose leveraging LLM checkpoints saved during pre-training, in conjunction with Bayesian optimization, to navigate the extensive search space and identify optimal merging coefficients. To address concerns about gradient mismatch, \citet{Lin2023MitigatingTA} introduce an adaptive method that assigns different combination ratios to various layers of the model, optimizing these combinations to balance human preference alignment and pre-training proficiency.

Other studies employ existing model merging techniques to create stronger LLMs for specific objectives. For example, \citet{splitandmerge} propose a disperse-then-merge framework, which first train multiple sub-models using different instruction-tuning data portions and then fuse them into a single LLM with weighted merging.

\paragraph{Enhancing RLHF} 
Model merging techniques also help improve the alignment of LLMs with human preferences.
\citet{Lin2023MitigatingTA} present an adaptive method where different combination ratios are assigned to various layers of the model, optimizing these combinations to balance human preference alignment and pretraining proficiency.
\citet{Rame2024WARMOT} propose fine-tuning multiple reward models and then averaging their parameters to create a superior reward model that aligns better with human preferences.
Similarly, \citet{fu2024disperse} use a weighted averaging approach to improve alignment in LLMs during the supervised fine-tuning (SFT), effectively reducing the impact of data bias. \citet{lu2024online} propose to use model merging technique to reduce alignment taxes during RLHF training for LLMs.

\subsection{Merging for Enhancing Multi-Task Capability (M-MTC)}
\label{sec:merging-multitask}

\begin{figure}[t]
    \centering
    \includegraphics[scale=0.37]{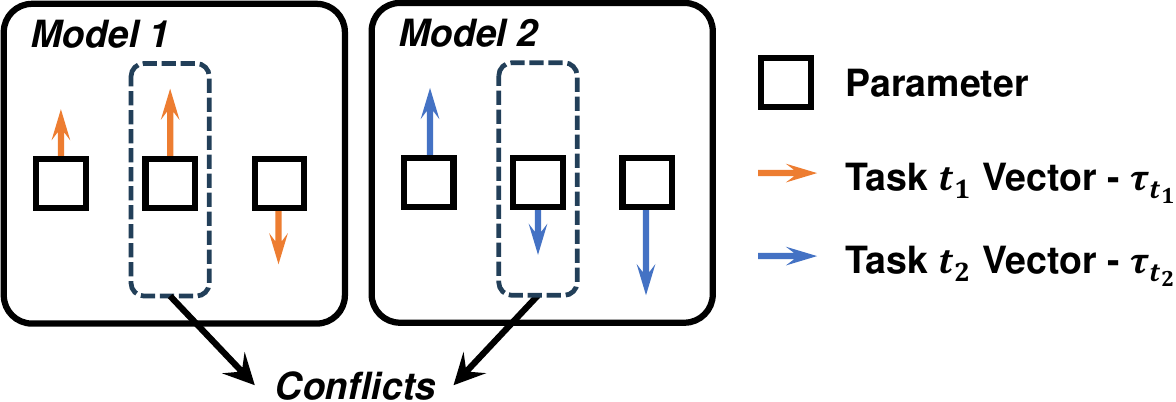}
    \caption{Illustration of the parameter conflicts. The direction denoting sign and length denoting magnitude of task vectors. The conflicts occur when the task vectors have opposite signs.}
    \label{fig.conflicts_illustration}
\end{figure}

Recently, some studies have attempted to merge models with different capability to construct a unified model with multi-task capability \cite{li2022branch}. Typically, these models are fine-tuned from the same pre-trained model but with different task-specific data, leading to divergence in their parameter spaces. Such divergence often reflects task-related information. Consequently, M-MTC methods aim to relieve the divergence and achieve a balanced merging of models with different capabilities, thereby producing a single model capable of handling multiple tasks. 

Early studies addressed the issue of divergence by using different merging coefficients for various models (\S \ref{sec:weight_average_multi_task}, Weight Average), while current research prefers to extract task properties from the divergence to achieve more flexible merging (\S \ref{sec:task_property}, Task Property). Furthermore, recent works have started to employ incremental learning techniques to enhance model merging performance (\S \ref{sec:incremental_Training}, Incremental Training). We separately introduce these methods in the following subsections.

\subsubsection{Methods based on Weighted Average}
\label{sec:weight_average_multi_task}

Some studies adopt weighted-average strategies (as introduced in \S \ref{sec:weight_average}) to adjust the importance of different models. \citet{jin2022dataless} propose RegMean to selectively integrate the linear layers of Transformer models while using simple averaging for other layers, thereby minimizing the divergence between the merged model and multiple models fine-tuned on various datasets. \citet{daheim2023model} advocate refining model merging using estimates derived from the Hessian matrix, facilitating more precise adjustments to model parameters. \citet{nathan2024fisher} combine Fisher weighted averaging with model pruning, achieving efficient model merging.

\subsubsection{Methods based on Task Property}
\label{sec:task_property}
Merging methods based on weighted average emphasize the importance of parameters but overlook their task-specific properties, leading to significant performance degradation in certain tasks. \citet{task_arithmetic} find that ``\textit{Simple Averaging suffers from a 10\% performance drop}''.
To address this issue, recent studies introduce a new paradigm known as the \textit{task vector}. \citet{task_arithmetic} define the task vector $\bm{\tau}_{t}$ as ``\textit{a vector specifies a direction in the parameter space of a pre-trained model, such that movement in that direction improves performance on the task}'', which can be formalized as:
\begin{gather}
\label{equ:task}
    \bm{\tau}_{t} = \bm{\theta}_{t}^{\text{ft}} - \bm{\theta}^{\text{pre}}
\end{gather}
where $\bm{\theta}_{t}^{\text{ft}}$ refers to the parameters fine-tuned with the specific data for task $t$, and $\bm{\theta}^{\text{pre}}$ refers to the original parameters of the pre-trained model.

Task vector can more effectively resolve parameter conflicts during the model merging.
As illustrated in Figure \ref{fig.conflicts_illustration}, using the pre-trained model as a reference, the variation in the direction of task vectors of the fine-tuned models indicates the presence of conflicts in the parameter space.
To address parameter conflicts problem, recent studies aim to exploring methods to mitigate conflicts and strike a balance between the different models. 
\textit{Parameter Conflict} methods resolve parameter conflicts at same position of parameters, while \textit{Fewer Parameter} methods identify and prune redundant parameters to reduce conflict. In addition, we introduce a tool that includes some methods in \textit{Toolkit}.

\paragraph{Resolving Parameter Conflicts}
\textsc{Task Arithmetic}~\cite{task_arithmetic} initially derives task vectors through arithmetic operations between fine-tuned and pre-trained parameters, as shown in equation (\ref{equ:task}). These task vectors are subsequently used to merge models for enhancing performance on target tasks.
To further address the issue of parameter conflicts in model merging, \textsc{Ties-Merging} \cite{ties} identifies two primary causes of interference: redundant parameter values and sign inconsistencies across models.
Building on these advancements, \textsc{AdaMerging}\cite{AdaMerging} reduces conflicts even further by considering the importance of different model parameters.
\textsc{MetaGPT}~\cite{metagpt} proposes efficient methods based on \textsc{AdaMerging} and \textsc{Task Arithmetic}.
\citet{evaluate_merge} propose a merging method for LLMs, which not only employ \textsc{Tie-Merging} for merging in parameter space, but also adopt evolutionary algorithms to optimize the data inference path inside the merge model. 
The above methods have been successfully adapted to LLMs.
\citet{Kim2024Prometheus2A} apply the above methods to fuse the LLMs obtain a stronger LLM evaluator. \citet{hammoud2024model} investigate the effects of above methods on LLM safety alignment.

Unlike the aforementioned task vector based methods to resolve conflicting parameters, \citet{zipit} propose \textsc{ZipIt} that retains similar parameters from another perspective.
\textsc{ZipIt} first identifies highly correlated parameters between different models. It then merges these parameters while retaining significantly different layers, thus improving the merging flexibility.

\paragraph{Pruning Redundant Parameters}
Given that conflicts may exist in the parameters of different models, another solution is to employ pruning techniques to reduce these conflicts before merging. Such methods further enhances the relevance of parameters to the task, and we introduce these methods separately.
\textsc{DARE}~\cite{mario}, a technique that efficiently reduces redundancy in fine-tuned language models by dropping and rescaling parameters.
\textsc{DELLA-merging}~\cite{dellamerging} further selects important parameters for fusion on the basis of DARE.
As domain-specific data and training techniques grow, the distinctions between fine-tuned models and their base models become more significant. 
However, DARE experiences significant performance drops, resulting in insufficient capability to process multiple domains effectively.
\textsc{DPPA}~\cite{dppa} presents a dual-stage pruning approach as Dynamic Pruning Partition Amplification (DPPA) for effectively merging divergent fine-tuned large language models across different domains.

\paragraph{Toolkit}
Recently, \citet{goddard2024arcee} have developed Arcee's MergeKit, an open-source toolkit that integrates various model merging methods, including Model Soups, DARE, and \textsc{TIES-Merging}. This toolkit significantly advances the application of model merging strategies in LLMs\footnote{\url{https://github.com/arcee-ai/mergekit}}.

\subsubsection{Methods based on Incremental Training}
\label{sec:incremental_Training}
The aforementioned methods still suffer from performance degradation. Therefore, several approaches involving incremental training have been proposed to restore their original performance.
Concrete TA/AM~\cite{Concrete} aims to find a shared low-dimensional subspace within the model parameter space to minimize task interference without significantly impacting performance. Surgery~\cite{surgery} introduces a representation surgery technique to mitigate representation bias in multi-task model fusion.

\section{Ensemble}
\label{sec:ensemble}
Ensemble learning is another effective collaboration strategy that differs from model merging methods by focusing on the combination of model outputs. Traditional techniques like Adaboost \cite{adaboost}, Bagging \cite{breiman1996bagging}, and Stacking \cite{wolpert1992stacked} have significantly advanced machine learning research. In the era of LLMs, ensemble learning continues to be crucial, enhancing the overall performance of various LLMs.

However, LLMs typically solve various tasks through text generation, resulting in more flexible and naturally expressive outputs. Therefore, traditional ensemble methods designed for classification tasks cannot be directly applied to LLMs. To address this issue, many studies explore specific ensemble methodologies tailored for various LLMs\footnote{It is worth noting that many studies explore self-ensemble approaches, such as self-consistency \cite{wang2023selfconsistency} and rationale augmentation \cite{wang2022rationale}. Our survey focuses on ensembles across different LLMs and does not cover these methods.}. Additionally, the benefits of ensemble learning have inspired research into various applications of these techniques. In the following sections, we will separately introduce LLM ensemble methodologies and applications in detail.

\subsection{LLM Ensemble Methodology}
\label{sec:ensemble-methodology}
For different inputs, LLM that performs best is not always the same, prompting extensive research into ensemble methods for LLMs. Unlike classification-based machine learning models, LLMs typically generate a sequence of tokens as output. This output is often discrete, making direct ensemble challenging. Additionally, structural differences between various LLMs result in vocabularies and output distributions that are difficult to unify, further complicating ensemble strategies \cite{xu2024bridging}. 

\begin{figure*}[t]
    \centering
    \includegraphics[scale=0.20]{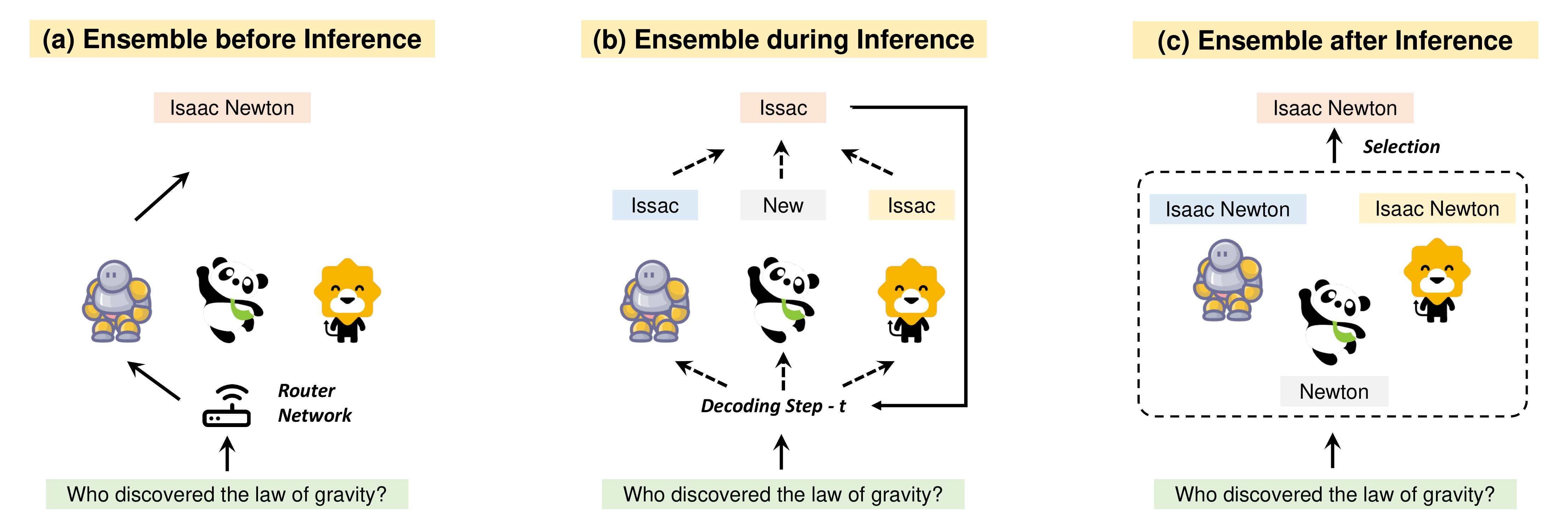}
    \caption{The illustrations of LLM \textit{ensemble} methods \textsc{Before} (a), \textsc{During} (b), \textsc{After} (c) inference.}
    \label{fig.ensemble_illustration}
\end{figure*}

Since ensemble generally occurs during the inference period, we categorize and introduce existing ensemble methods employed \textsc{Before}, \textsc{During}, and \textsc{After} the inference period. As illustrated in Figure \ref{fig.ensemble_illustration}, ensemble methods \textsc{Before} inference select the most suitable LLM for different input examples, ensemble methods \textsc{During} inference combine outputs at each decoding step, and ensemble methods \textsc{After} inference aim to select the best response from several outputs generated by various LLMs.

\subsubsection{Ensemble \textsc{Before} Inference}
\label{sec:ensemble-methodology-before}

Such methods aim to select the best LLM for specific examples before inference. Similar but different from various Mixture-of-Expert (MOE) approaches \cite{jacobs1991adaptive, collobert2003scaling, eigen2013learning, fedus2022switch, jiang2024mixtral}, which learn sparse networks from scratch, ensemble \textsc{Before} Inference focuses on training external routers \cite{rosenbaum2017routing} for several pre-trained LLMs to achieve optimal LLM selection.

\citet{shnitzer2023large} take the lead to explore the feasibility and limitations of learning routers by using various benchmark datasets. \citet{lu2023routing} introduce \textsc{Zooter}, a system that first employs a reward model to calculate scores for query-output pairs using the training set. These scores are then utilized to train a router using the knowledge distillation strategy, allowing it to select the optimal LLM based solely on input queries. \citet{ding2024hybrid} employ a router that assigns queries to either a small model or LLMs based on the predicted query difficulty and the required quality level, significantly reducing inference costs. \citet{srivatsa2024harnessing} investigate the feasibility of classifier-based and clustering-based routing methods for LLMs. Inspired by self-play in reinforcement learning, \citet{mohammadshahi2024leeroo} train the router by recycling the self-produced triplets, (\textit{query}, \textit{response}, \textit{score}). Unlike previous studies, \citet{lu2024blending} integrate multiple Chat LLMs by randomly selecting an LLM at each turn in the dialogue, rather than learning a router. To effectively evaluate the router capability and limitations, \citet{hu2024routerbench} propose a new benchmark, \textsc{RouterBench}, mainly focusing on performance and
economic cost.


\subsubsection{Ensemble \textsc{During} Inference}
\label{sec:ensemble-methodology-during}

During the inference period, LLMs generate tokens auto-regressively. This process often results in early errors compounding over time, causing subsequent tokens to deviate from the intended meaning \cite{ranzato2016sequence} and leading to hallucinations \cite{zhang2023language}.

To address this problem, some studies perform ensemble LLMs at each decoding step. \citet{li2024purifying} combine untrusted LLMs with a benign smaller LLM by weighted-averaging the output distributions, mitigating issues such as copyright infringement, data poisoning, and privacy violations. \citet{hoang2023fly} interpolate the output distributions from the machine translation model and the LLM, boosting translation performance. \citet{wang2024fusing} formulate the \textit{Frugal Fusion of Experts} problem and proposes an efficient fusion method by addressing it as a graph shortest path problem. These methods require the ensemble to occur among LLMs which must at least have the same vocabulary. This restriction ensures that the output distributions are aligned and can be interpolated effectively.

However, most open-source LLMs are heterogeneous and have different vocabularies, hindering direct ensembling. To address this issue, \citet{fu2023specializing} employ dynamic programming to recursively minimize the total cost measured by the exact match scores of editing one sequence of tokens to match another. To further enhance the success rate of token alignments, \citet{wan2024knowledge} replace the exact match constraint with a minimum edit distance strategy. \citet{mavromatis2024pack} follow the above token alignments, proposing the use of perplexity to compute the coefficients for the outputs of diverse LLMs during ensemble. \citet{xu2024bridging}, \citet{huang2024enabling}, and \cite{yu2024breaking} consider overlapping tokens as anchors to project the output distribution produced by heterogeneous LLMs into the same space. Specifically, \citet{xu2024bridging} propose to directly learn the projection matrices between different vocabularies using the anchors as bridges, while \citet{huang2024enabling} and \citet{yu2024breaking} calculate the relative representations from anchors to different vocabularies, thereby indirectly achieving the vocabulary projection.

\subsubsection{Ensemble \textsc{After} Inference}
\label{sec:ensemble-methodology-after}

The final LLM ensemble methods combine the generated outputs \textsc{After} the inference period.

One approach to achieving the LLM ensemble involves building LLM cascades to reduce the inference cost associated with using large LLMs exclusively.
\citet{chen2023frugalgpt} employ a sequence of LLMs ranked by the number of parameters to generate outputs, halting the process and returning results once a preceding smaller LLM produces outputs of sufficient quality. \citet{yue2024large} propose verifying the correctness of answers generated by a smaller LLM first and utilizing LLMs to solve the problem if the initial answer is incorrect.

Another line of research focuses on selecting the best candidate from several produced by various LLMs.
\citet{lee2023ensemble} select the best instruction from several candidates produced by LLMs for instruction-tuning data construction. \citet{jiang2023llm} explore various unsupervised metrics for selection, including BERTScore \cite{Zhang2020BERTScore}, BLEURT \cite{sellam2020bleurt}, BARTScore \cite{yuan2021bartscore}, and ChatGPT scores. However, they find that the effectiveness of the selection is constrained by the quality of candidate pools. To address this problem, \citet{jiang2023llm} adopt an additional fusion model to generate the final output using the top-ranked candidates as inputs.

\begin{table*}[htpb]
\centering
\begin{tabular}{cccc}
\hline
\textbf{Inference Stage} & \textbf{Inference Speed} & \multicolumn{1}{c}{\textbf{Ensemble Granularity}} & \textbf{Limitations \& Challenge}           \\ \hline
\textsc{Before}                   & $\searrow $              & \textit{example-level}                            & Accuracy of Routers          \\
\textsc{During}                   & $\searrow \searrow$    & \textit{token-level}                              & Heterogeneous Architectures  \\
\textsc{After}                    & $\searrow \searrow \searrow$     & \textit{example-level}                            & Accuracy of Candidate Pools \\ \hline
\end{tabular}
\caption{The characteristic of LLM ensemble methods employed at different inference stages.}
\label{tab.characteristic}
\end{table*}

\subsubsection{Discussion about LLM Ensemble Methods}
\label{sec:ensemble_discussion}
The methods described above focus on ensemble techniques at different stages of inference, each with its own strengths and weaknesses. We discuss them in terms of \textit{inference speed}, \textit{ensemble granularity}, and \textit{limitations}.

\paragraph{Inference Speed}
\textit{Almost all ensemble methods decrease inference speed}. Ensemble methods \textsc{Before} inference slightly slow down the process, as they involve selecting the optimal LLM via additional routers. Ensemble methods \textsc{During} inference require each LLM to perform forward computation for test examples. If we have $k$ LLMs, the inference speed will slow down by a factor of $k$, although this can be mitigated by using $k$ times the number of GPUs. Ensemble methods \textsc{After} inference not only require $k$ times the computational cost but also spend additional time on output selection or fusion \cite{jiang2023llm}, resulting in the lowest inference speed.

\paragraph{Ensemble Granularity}
The aforementioned ensemble methods operate at different levels of granularity. Ensemble methods \textsc{Before} and \textsc{After} inference typically work at the example level, providing a coarse-grained ensemble. Ensemble methods \textsc{During} inference perform ensemble at the token level, offering a fine-grained approach. Since preceding tokens often influence succeeding tokens, this fine-grained ensemble can reduce exposure bias \cite{ranzato2015sequence, xu2020rethinking} and decrease hallucination in LLMs, thus holding better potential for performance enhancement.

\paragraph{Limitations}
Each category of ensemble methods have specific limitations, as shown in Table \ref{tab.characteristic}. Although ensemble methods \textsc{Before} inference offer better speed, they require additional training of the router. The data used for router training can limit the generalization and performance of these ensemble methods \cite{shnitzer2023large}. Ensemble methods \textsc{During} inference are typically constrained by the heterogeneous architecture of LLMs. For example, differences in vocabularies can lead to non-corresponding output distributions, hindering direct ensemble \cite{huang2024enabling, yu2024breaking}. Ensemble methods \textsc{After} inference often require multiple LLMs to generate several candidates, construct a candidate pool, and then select or reorganize the final output. Thus, the accuracy of the candidate pool \cite{jiang2023llm} and the selection strategy are the main limitations.

\subsection{LLM Ensemble Application}
\label{sec:ensemble-application}

In addition to methodological research, many studies apply LLM ensembles to specific applications. This is due to the capability of ensemble learning ability on specific tasks, domains and strong calibration. We categorize these related studies based on their objectives.

\paragraph{LLM Ensemble for Specific Tasks or Domains}

Ensemble learning for LLMs are typically employed for specific tasks. \citet{si2023getting} improve LLM reasoning performance by training a classifier to select the best answer generated by various reasoning experts. \citet{gundabathula2024promptmind} employ LLM ensemble for SQL generation. Some studies employ LLM ensemble for medical tasks. \citet{gundabathula2024promptmind} adopt LLM ensemble to enhance the clinical text error detection and correction. \citet{oniani2023large} and \citet{barabucci2024combining} utilize majority voting and average weighting, respectively, to select the most frequent candidate disease.  





\paragraph{LLM Ensemble for Overestimation Mitigation in RLHF} Ensemble learning can alleviate the poor calibration and unreliable prediction problems of LLMs. Therefore, \citet{eisenstein2023helping}, \citet{coste2024reward} and \citet{rame2024warm} perform ensemble with multiple reward LLMs to mitigate the overoptimization problem in RLHF. Considering that fine-tuning reward models based on LLMs can be computationally expensive, \citet{zhang2024improving} and \citet{zhai2023uncertainty} respectively employ light-weight LoRA \cite{hu2022lora} to adapt the LLM to multiple reward models. \citet{ahmed2024scalable} propose using a shared LLM but separate linear heads for reward ensemble. These methods successfully reduce the overestimation in RLHF and improve the alignment performance..



\section{Cooperation}
\label{sec:cooperation}
In the era of LLMs, collaborative strategies extend beyond mere merging or ensemble. Increasingly, studies are focusing on broader approaches to solving various problems or specific tasks through cooperation between LLMs. In the following sections, we introduce the cooperation strategies based on their objectives: \textit{efficient computation} (\S \ref{sec:cooperate-accelerate}), \textit{knowledge transferring} (\S \ref{sec:cooperate-transferring}), \textit{compensatory cooperation} (\S \ref{sec:cooperate-compensatory}), \textit{federated cooperation} (\S \ref{sec:cooperate-federated}).

\subsection{Efficient Computation}
\label{sec:cooperate-accelerate}

As LLMs grow in scale, the computational resources required for their inference increase significantly. Consequently, accelerating model inference has become an urgent necessity. Smaller LLMs play a crucial role in accelerating larger LLMs due to their lightweight nature \citep{Miao2023SpecInferAG}. Research on model acceleration through cooperation can be divided into two main categories: Input Compression (\S \ref{sec:cooperate-accelerate-predecoding}) and Speculative Decoding (\S \ref{sec:cooperate-accelerate-decoding}). Input Compression achieves efficient computation by using smaller LLMs to compress inputs, thereby reducing the context length. Speculative Decoding involves leveraging smaller LLMs to draft multiple tokens speculatively, with larger LLMs verifying these draft tokens in parallel.

\subsubsection{Input Compression}
\label{sec:cooperate-accelerate-predecoding}

\begin{figure}[h]
    \centering
    \includegraphics[width=0.9\linewidth,scale=1.00]{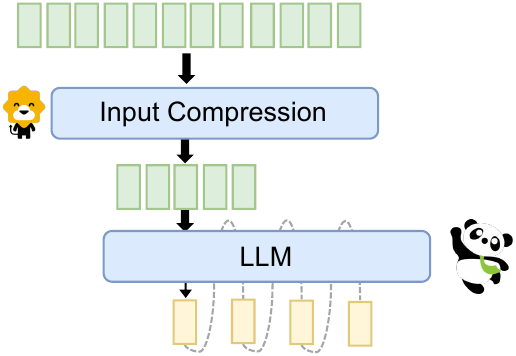}
    \caption{LLMs cooperate with compression module for input compression.}
    \label{fig:example-distribution}
\end{figure}

Input compression aims to use small models to compress input prompts or long text content. When processing long input sequences, the self-attention mechanism of LLMs results in significant increases in computational cost and memory usage due to its quadratic time complexity \cite{xu2024survey}. Input compression reduces computational costs, memory usage, and access costs by shortening input lengths and reducing the number of tokens the model needs to handle during the pre-fill phase, thereby improving memory efficiency \cite{wan2024efficient}. Additionally, input compression accelerates token generation in autoregressive decoding by shortening input sequence lengths, thus enhancing inference efficiency.

\citet{Zhou2024ASO} propose a review on efficient inference, where input compression being a significant part of the review. Input compression mainly focuses on \textit{prompt pruning}, \textit{prompt summarization}, and \textit{soft prompt compression}. The core idea of prompt pruning is to delete unimportant tokens, sentences, or documents in the input prompt based on predefined or learnable importance metrics. Prompt summarization aims to compress the original prompt into a shorter summary while retaining similar semantic information. Soft prompt compression involves designing a soft prompt which is a sequence of learnable continuous tokens. It is much shorter than the original prompt and will be utilized as the input for LLMs. Among the above-mentioned methods, they usually need to cooperate with a compression model, such as a summarization model, to achieve the input compression. 

\paragraph{Prompt Pruning} The core idea behind the prompt pruning is to remove unimportant tokens, sentences, or documents, where the cooperation model targets at prompt pruning. \citet{ali2024promptsaw} utilize graph-construct model to obtain a graph from the textual information in the prompt and extracting key information elements from the graph to obtain the compressed prompt. \citet{pan2024llmlingua2} introduce the LLMLingua-2 distillation method, which classifies each input token using a Transformer encoder. Then it preserves the top $N$ tokens with the highest classifying probability, capturing all key information for prompt compression. 
\cite{huang2024fewermoreboostingllm} propose a coarse-to-fine pruner that initially identifies crucial CoT examples from a large batch and then further prunes the unimportant tokens.

\paragraph{Prompt Summarization} The core idea of prompt summarization is to condense the original prompt into a shorter summary which preserves the same semantic information, where the cooperation model is an extractive or abstractive summarization model. \citet{liu-etal-2023-tcra} propose two compression methods. The first method trains a summarization model to compress the context. The second method further reduces the number of tokens by deleting words with less semantic impact. RECOMP \citep{xu2023recompimprovingretrievalaugmentedlms} introduces an abstractive compressor that takes a question and retrieved documents as input to produce a concise summary. Then the LLM generate the answer based on the concise summary. SemanticCompression \citep{fei2023extendingcontextwindowlarge} proposes a method that breaks text into sentences, groups sentences by topic, and then summarizes each group with pre-trained model. Finally, the LLM generate the response based on all the groups. To address the different tasks, \citet{li2024pctoolkit} propose PCToolkit, which consists compressors that are developed with different targets.

\paragraph{Soft Prompt Compression} Soft prompt compression involves using virtual tokens to assist in prompt compression, where the cooperation model is an text feature extractor. For example, SelfCP \cite{gao2024selfcp} uses a frozen LLM as an encoder and decoder, inserting special tokens into the prompt to generate virtual tokens, thereby achieving prompt compression and response generation collaboration.  \citep{gao2024unifying} projects candidate demonstrations into virtual tokens via a LLM during in-context learning, then it selects appropriate demonstrations based on semantic similarity. Finally, it generates the response using a frozen LLM.

In general, we can observe that from \textit{prompt pruning} to \textit{prompt summarization} to \textit{soft prompt compression}, the compression ratios increase which results in higher efficiency. However, the corresponding information loss is also greater.
\subsubsection{Speculative Decoding}
\label{sec:cooperate-accelerate-decoding}

\begin{figure}[h]
    \centering
    \includegraphics[width=0.9\linewidth,scale=1.00]{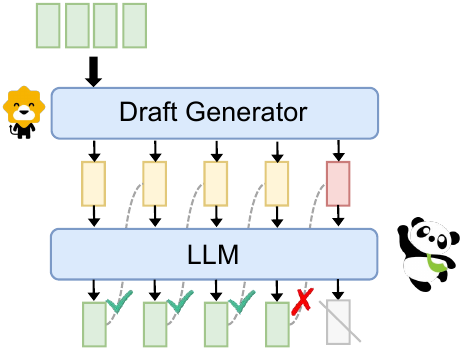}
    \caption{LLMs cooperate with draft generator for speculative decoding.}
    \label{fig:example-distribution}
\end{figure}

The primary focus of decoding acceleration is to cooperate smaller models to generate drafts, which are then verified by larger models to expedite inference \cite{stern2018blockwise}. This approach is currently referred to as speculative decoding \cite{leviathan2023fast, xia-etal-2023-speculative}. \citet{xia2024unlocking} summarize various drafting strategies, including Independent Drafting and Self-Drafting. The drafted tokens are then verified in parallel by the target LLM, with verification methods such as Greedy Decoding \citep{gu2017trainablegreedydecodingneural} and Nucleus Sampling \citep{holtzman2020curiouscaseneuraltext}. The acceleration effect of speculative decoding largely depends on the acceptance rate of drafted tokens at each step. This acceptance rate is influenced by several factors, including the quality of the drafts, the verification criteria, and the behavioral consistency between the drafter and the target LLM.

For instance, \citet{Ou2024LosslessAO} propose the use of an adaptive \textit{N}-gram model based on the current context for the rapid drafting phase, followed by a verification phase where the original LLM evaluates and confirms the proposed tokens.
In addition to using draft models to reduce the number of actual model predictions, some studies have introduced early stopping mechanisms. This method terminates the prediction process early based on specific conditions, thereby saving computational resources. For example, \citet{huang2024specdec++} and \citet{liu2024kangaroo} propose using probabilistic predictions to assess the acceptance rate of draft model's hypotheses, deciding whether to continue generating more drafts based on predefined thresholds. Furthermore, \citet{liu2024speculative} incorporate an early stopping mechanism after the initial \textit{N} layers of the model during draft hypothesis generation.

\subsection{Knowledge Transfer via Cooperation}
\label{sec:cooperate-transferring}

LLMs typically encompass various capabilities. However, due to the difficulty in obtaining training data and the high training costs, directly transferring knowledge or capabilities from one LLM to another has attracted significant attention. Considering that the output probabilities of LLMs often contain the embedded knowledge of the models, recent methods have primarily focused on transferring knowledge between LLMs. For example, \citet{wan2024knowledge, wan2024fusechat} transfer knowledge from multiple LLMs to a target model via continued training with knowledge distillation \cite{hinton2015distilling}. Nonetheless, most of the recently proposed methods focus on cooperation at the inference stage without involving training and can be categorized based on their objectives: \textit{Mitigating Incorrect Knowledge}, \textit{Strengthening Correct Knowledge}, and \textit{Supplying New Knowledge}.

\subsubsection{Mitigating Incorrect Knowledge}
\label{sec:cooperate-transfer-wiping_hallucination}
Hallucination \cite{rawte2023survey, huang2023survey} and bias \cite{wang2024decodingtrust} are common issues in LLMs. \citet{ji2023survey} argue that LLMs trained with the next-word prediction objective store superficial patterns instead of recognizing real-world knowledge extracted from the training corpora. Therefore, many studies aim to explore cooperation methods for removing incorrect knowledge existed in the output logits.

\begin{figure}[h]
    \centering
    \includegraphics[width=1.0\linewidth,scale=1.00]{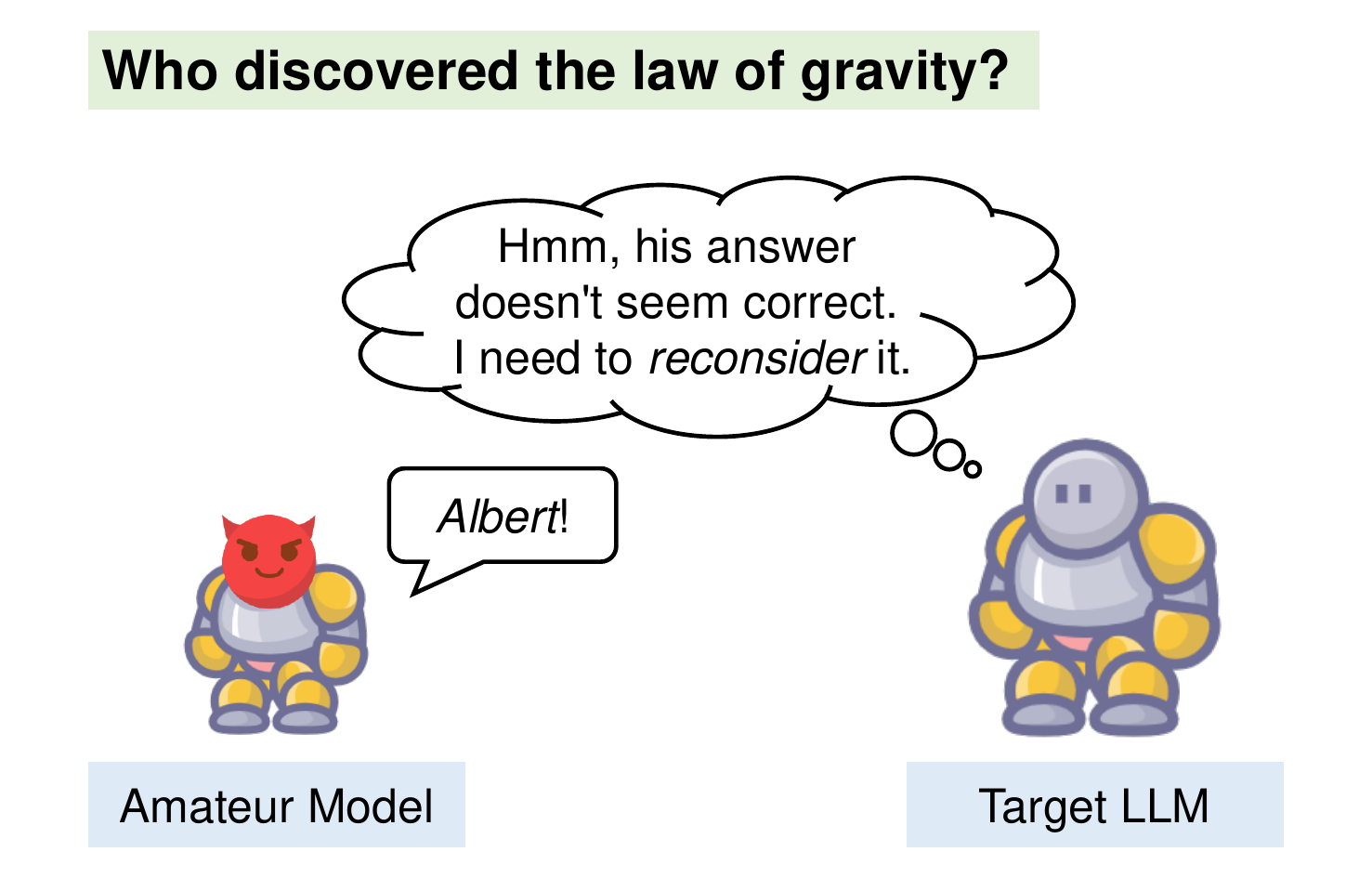}
    \caption{Illustration of core idea of \textsc{Cd}. The amateur model is more prone to errors, thus the target LLMs needs to reconsider the answer accordingly.}
    \label{fig:illustration_mitigation}
\end{figure}

Drawing on the observation that the shortcomings of larger LLMs are even more pronounced in smaller ones, \citet{li2023contrastive} propose Contrastive Decoding (\textsc{Cd}). \textsc{Cd} encourages token $y_{i}$ selection from the delta distribution between LLMs and their corresponding weaker, amateur models, thereby eliminating incorrect knowledge from output distributions:
\begin{gather}
    y_{i} \sim \underbrace{\log p_{\text{LLM}}(y_{i} | y_{<i})}_{\text{vanilla distribution}} - \underbrace{\log p_{\text{AMA}}(y_{i} | y_{<i})}_{\text{amateur distribution}}
\end{gather}

This work has significantly inspired subsequent studies. Some studies employ \textsc{Cd} to enhance reasoning capabilities. \citet{o2023contrastive} demonstrate that \textsc{Cd} improves reasoning for LLMs by preventing certain abstract reasoning errors. \citet{phan2024distillation} utilize distillation techniques to obtain amateur models and perform contrastive decoding to enhance reasoning capabilities. Additionally, some studies use \textsc{Cd} for trustworthy generation. \citet{waldendorf2024contrastive} experiment with \textsc{Cd} in various amateur models for machine translation, showing that \textsc{Cd} reduces hallucinations in large multilingual machine translation models. \citet{liu2021dexperts} employ \textsc{Cd} for language detoxification and sentiment-controlled generation. \citet{zhang2023alleviating} and \citet{niu2024parameter} induce hallucinations or toxic content into LLMs by fine-tuning with non-factual or toxic examples and then use these models as amateurs for contrastive decoding. \citet{qu2024unsupervised} leverage counterfactual \textsc{Cd} for distractor generation. Some studies explore diverse contrastive strategies. \citet{shi2024unchosen} observe that output distributions from an MoE model using different routing strategies differ substantially, thus they utilize unchosen experts as the amateur model to achieve contrastive decoding. \citet{yuan2023speculative} combine \textsc{Cd} with speculative decoding to achieve both acceleration in decoding and quality improvements.

Although \textsc{Cd} can help mitigate incorrect knowledge during inference, LLMs and the corresponding amateur models typically belong to the same family and require aligned output distributions.

\subsubsection{Strengthening Correct Knowledge}
\label{sec:cooperate-transfer-Verifying}
Beside mitigating hallucination or bias from the output distributions, another line of research focuses on enhancing the faithfulness of decoding outputs to the input or instructions through LLMs cooperation.  This typically involves using additional models to strengthen the correct knowledge - increasing the likelihood of potentially correct outputs. This line of research can be traced back to studies on attribute-controlled text generation, which encourage language models to output tokens that effectively predict the input attributes, thereby improving faithfulness.Notable examples include \textsc{Fudge} \cite{yang-klein-2021-fudge} and \textsc{GeDi} \cite{krause-etal-2021-gedi-generative}, which are based on Bayesian factorization:
\begin{gather}
    p(y_{i}|y_{<i}, c) \propto p(y_{i} | y_{<i}) \cdot p(c|y_{i}, y_{<i}) \\
    \label{equ.verification}
    y_{i} \sim \underbrace{\log p_{\text{LLM}}(y_{i} | y_{<i})}_{\text{vanilla distribution}} + \underbrace{\log p_{\textsc{Ver}}(c|y_{i}, y_{<i})}_{\text{verification}}
\end{gather}
where $c$ is the input attribute and $\log p(c|y_{i}, y_{<i})$ is derived from additionally learned attribute predictors, namely \textit{verifier}. It should be noted that any suitable model, including the LLM itself, can serve as the \textit{verifier}.


\begin{figure}[h]
    \centering
    \includegraphics[width=1.0\linewidth,scale=1.00]{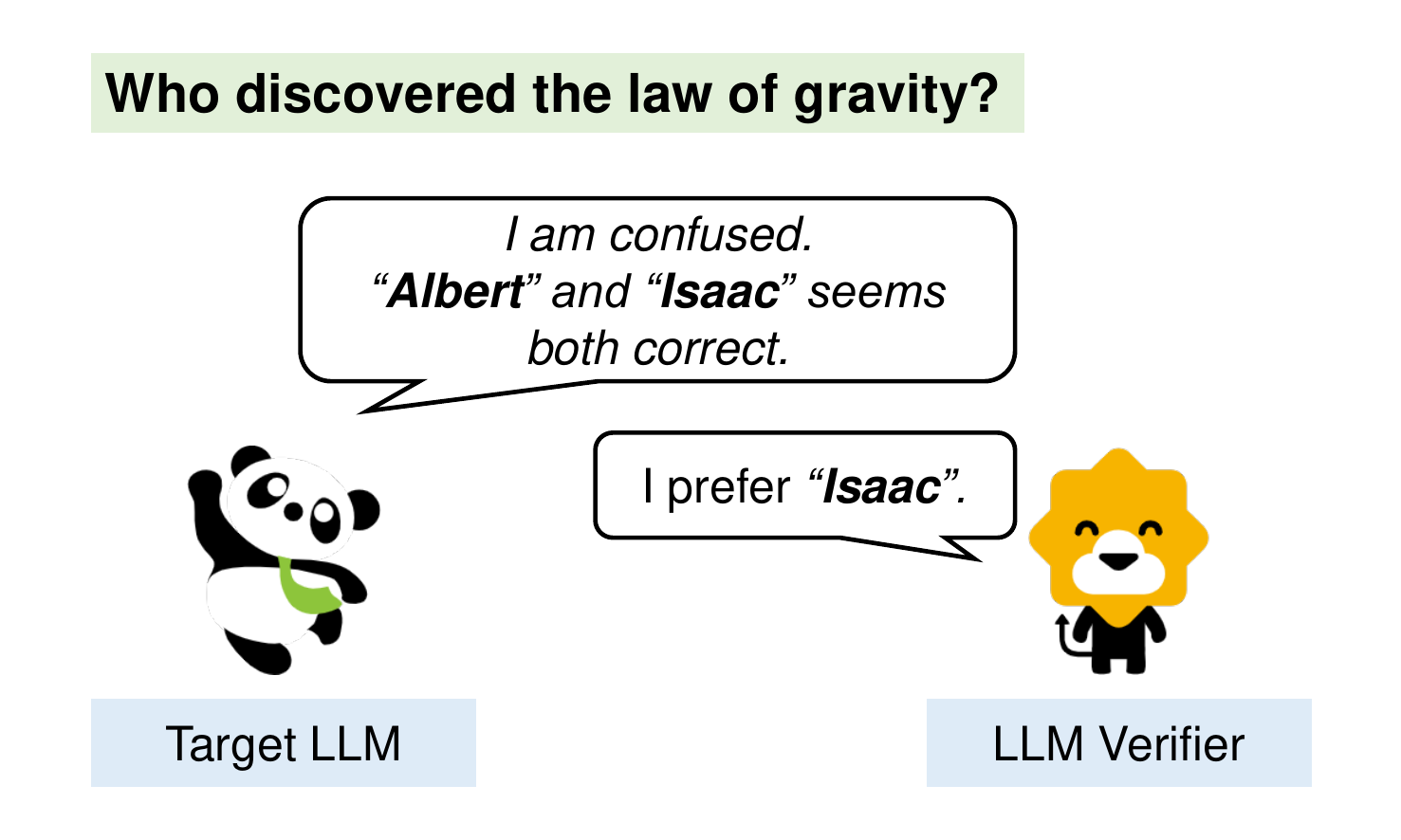}
    \caption{Illustration of core idea of \textit{verification} based methods. The LLM \textit{verifier} needs to check and assist selecting the correct output from candidates.}
    \label{fig:illustration_verifying}
\end{figure}

Recently, \citet{deng-raffel-2023-reward} employ an auxiliary reward model as the verifier to encourage LLMs to generate text that has certain properties. \citet{tu2023unlocking} verify each decoding position according to equation (\ref{equ.verification}) in language generation tasks, enhancing the faithfulness of LLMs. To achieve accurate verification with sufficient information, \citet{lu2024diver} introduce \textsc{Diver}, which employs dynamic token spans that can potentially be generated to calculate point-wise mutual information (PMI) for verification. Additionally, they demonstrate that using smaller LLMs for verification can alleviate the decrease in inference speed without significantly affecting performance.



\subsubsection{Supplying New Knowledge}
\label{sec:cooperate-transfer-Supplying}
Recently, some studies have observed that the variation of output logits reflect the variation of LLMs capabilities. Inspired by this, they propose to tune larger LLMs without training. Instead, they modify the output logits to provide new capabilities extracted smaller LLMs:
\begin{equation}
\label{equ.supply_capability}
    \begin{split}
    & \log p_{\textsc{Large-C}}({y_{i}|y_{<i}}) \propto \underbrace{\log p_{\textsc{Large}}({y_{i}|y_{<i}})}_{\text{vanilla distribution}} \\
    & + \underbrace{\left[ \log p_{\textsc{Small-C}}({y_{i}|y_{<i}}) - \log p_{\textsc{Small}}({y_{i}|y_{<i}}) \right]}_{\text{capability reflection}}
    \end{split}
\end{equation}
where \textsc{Large-C} and \textsc{Small-C} refers to models with specific capability while \textsc{Large} and \textsc{Small} refers to the original models.

\citet{ormazabal-etal-2023-comblm} and \citet{liu2024tuning} separately propose emulated tuning and proxy tuning, extracting \textit{Chat} capability from smaller LLMs and integrate the capability into larger LLMs with the aforementioned equation (\ref{equ.supply_capability}). \citet{zhao2024weak} employ safe and unsafe smaller models to adversarially modify a significantly larger safe model’s decoding probabilities, achieving jailbreaking for LLMs. 

The above methods require that both small and large LLMs share the same vocabulary. To address this issue, \citet{zhou2024weak} introduce \textit{weak-to-strong search}, utilizing the log-likelihood difference between small tuned and untuned LLMs as rewards to guide the decoding of larger LLMs with tree search, mitigating the need for shared vocabularies.

\label{sec:transferring-training}

\subsection{Compensatory Cooperation}
\label{sec:cooperate-compensatory}

In practical applications, large models still face some uncontrollable issues due to their lack of interpretability \citep{10.1145/3639372}. Therefore, it is necessary to introduce additional controllers to compensate for the deficiencies of LLMs. Depending on the desired properties, controllers can be served as: 1) detector, 2) retriever, etc..

\subsubsection{Detector}
\label{sec:cooperate-compensatory-detector}
\begin{figure}[h]
    \centering
    \includegraphics[width=0.9\linewidth,scale=1.00]{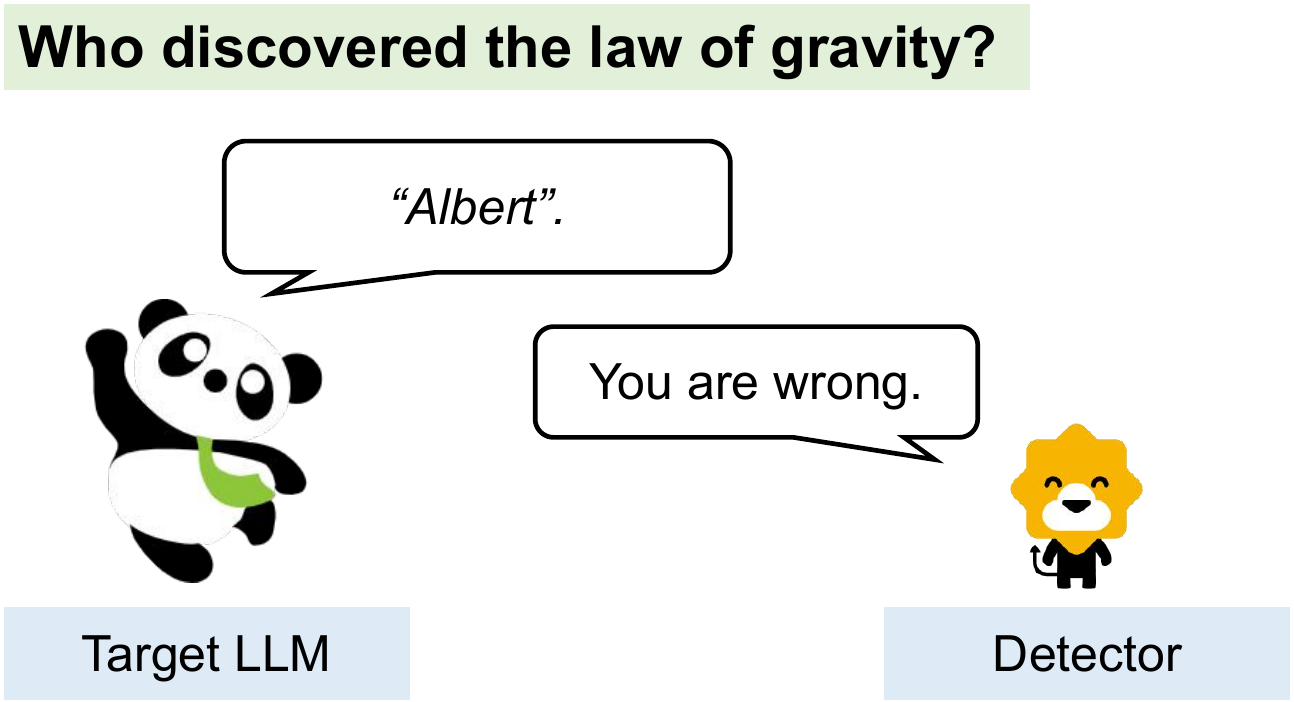}
    \caption{LLMs cooperate with detector.}
    \label{fig:example-distribution}
\end{figure}
LLMs may generate incorrect responses or hallucinations due to a lack of relevant knowledge. Therefore, detecting hallucinations in LLMs is crucial for ensuring the reliability and credibility of generated content. We introduce related work from two types of hallucinations: factual hallucinations and faithfulness hallucinations.

\paragraph{Factual Hallucinations.} Factuality hallucination emphasizes the discrepancy between generated content and verifiable real-world facts \citep{bai2024hallucination}. Current mainstream methods for detecting factual hallucinations can be divided into: 1) retrieve external facts: identifying factual inaccuracies by comparing the model-generated content against external knowledge sources \citep{min2023factscore, Tang2024MiniCheckEF}. 2) uncertainty estimation: retrieving external facts introduces additional cost. Therefore, several methods rely on internal knowledge to alleviate hallucinations \citep{Chen2024INSIDELI, Tsai2024EfficientNU}. Among them, the cooperation studies of LLMs mainly falls into the first category.

Some studies directly utilizes additional models as detectors to identify fact errors in the output of LLMs. RARR \citep{Gao2022AttributedTG} leverages a NLI model to evaluate the factulity between the generated sentences and evidence. Consider that some evidence may not fully support a claim, thus Factcheck \citep{Wang2023FactcheckBenchFE} identifies the stance of evidences and gathers the relevant evidence, then it uses RARR to determine whether the gathered evidence supports the original text. Additionally, to detect the factuality of the model regarding complex multi-step problems, PROGRAMFC \citep{Pan2023FactCheckingCC} decomposes complex claims into simpler sub-tasks that can be solved using question-answer module, NLI module, in-context learning module. However, single-pass verification cannot fully utilize the experience gained from similar cases and may repeat past errors. Therefore, SALAM \citep{wang-li-2023-learning} exploits a auxiliary model to assist the main LLM in learning from mistakes through interactive cooperation model.

Other studies exploits auxiliary models to collect relevant evidence for verification \citep{chen2023complex, Huo2023RetrievingSE, Ousidhoum2022VarifocalQG, Yao2022EndtoEndMF}. \citet{Huo2023RetrievingSE} adopt information retrieval model to obtain related passages for factual detecting. The retrieved documents contain up to several thousand words, which becomes cumbersome for both humans and models to make a judgment based on them. Therefore, \citet{chen2023complex} compress retrieved text with an summarization model and obtains relevant claims, which is provided for verifying the fact of LLMs output.

\paragraph{Faithfulness Hallucinations.} Faithfulness hallucination refers to the divergence of generated content from user instructions or the context provided by the input \citep{bai2024hallucination}. The current mainstream methods for detecting faithfulness hallucinations can be divided into: 1) rule-based: which measures the overlap of pivotal facts between the generated content and the source content \citep{Nan2021EntitylevelFC, Goodrich2019AssessingTF}; 2) classifier-based, leveraging trained classifiers to obtain the degree of entailment between the generated and the source content \citep{Falke2019RankingGS,Mishra2021LookingBS}; 3) QA-driven, employing question-answering systems to verify the consistency between the generated and the source content; 4) uncertainty estimation, which evaluates faithfulness by assessing the model’s confidence in its generated outputs; 5) prompting-based, where LLMs are prompted to function as evaluators, assessing the faithfulness of generated content through specific prompting strategies. Among these, the collaborative studies of LLMs mainly falls into categories 2) and 3). Their difference mainly lies in the use of different cooperation models for hallucination detection.

The classifier-based methods employ NLI models to filter out generated sentences that is not supported by the fact of evidence document \citep{Chen2023UnderstandingRA, Wang2024LLMsKW}. \citet{Chen2023UnderstandingRA} exploit an NLI model to make entailment decisions for each document sentence and answer sentence pair,  then aggregates the results by taking the maximum value over all the pairs. To avoid obtaining hallucinated information in the processs of retrieval, \citet{Wang2024LLMsKW} utilize an NLI model to determine whether the retrived passages indeed entail the target information. The QA-driven methods apply QA model to obtain the faithfulness of the generation. \citet{qiu-etal-2023-detecting} formulate faithfulness evaluation as binary classification problem. Then it develop a classifier to distill knowledge from QAFactEval \citep{fabbri-etal-2022-qafacteval}. 

In brief, by collaborating with a different detectors, LLMs can often verify or correct some errors, thereby alleviating the issue of uncontrolled content generation.

\subsubsection{Retriever}
\label{sec:cooperate-compensatory-retriever}
\begin{figure}[h]
    \centering
    \includegraphics[width=0.9\linewidth,scale=1.00]{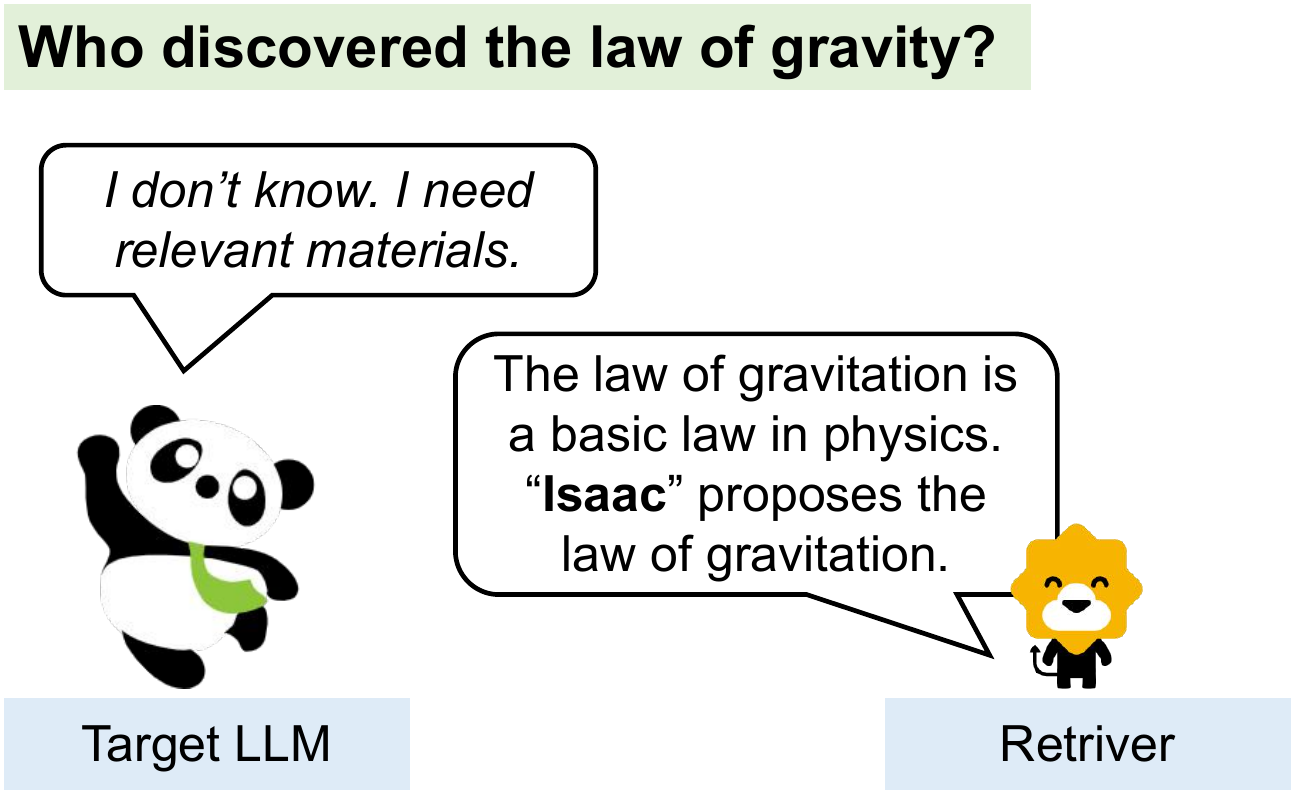}
    \caption{LLMs cooperate with retriever.}
    \label{fig:example-distribution}
\end{figure}

The information possessed by LLMs is limited. To extend the knowledge of large models, retrieval augmentation generation (RAG) \citep{gao2024retrievalaugmentedgenerationlargelanguage} aims to use external data sources to assist text generation. Collaborative models serve as bridges to external knowledge, assisting in the retrieval of information. The cooperation models for retrieving different types of data vary significantly, including: 1) unstructured data; 2) structured data.

For unstructured data, such as text, some studies retrieve world knowledge to expand the corpus of the LLMs. \citet{Izacard2022FewshotLW} start by retrieving the top-k relevant documents from a large corpus of text with Contriever \citep{Izacard2021UnsupervisedDI}. Then, these documents are fed to the LLM, along with the query, which in turns generates the output. Besides, it is inconvenient to train language models with special cross attention mechanisms to encode the retrieved text. Thus REPLUG \citep{Ma2023QueryRF} treats the LLMs as a black box and augments it with a tuneable retrieval model. \citet{ma-etal-2023-query} note that there is inevitably a gap between the input text and the needed knowledge in retrieval. Therefore, to better align the query to the retriver, it adopts a small language model as a trainable rewriter to cater to the black-box LLM reader. Howerver, query rewriting lacks the utilization of effective and general signals. Therefore, RaFe \citep{Mao2024RaFeRF} first trains an initial query rewriting model by standard supervised fine-tuning, Subsequently, it utilizes the ranking scores from the reranker to conduct feedback training on the query rewriting model.

For structured data, such as knowledge graph and SQL, it requires additional effort to convert structured data to text \citep{li2024chainofknowledge, su2024semistructured, wang2023knowledgpt, he2024gretriever}. Therefore, the retriever needs to have the ability to convert between natural language and structured language. For knowledge graph, the retriever is an entity linking model to retrieve relative structured knowledge \citep{su2024semistructured,wang2023knowledgpt}. Additionally, for SQL knowledge, the retriever is a query generator that allows the generation of queries for various types of query languages, including SPARQL, SQL, and natural sentences \citep{li2024chainofknowledge, zha2023tablegpt}.


\subsection{Federated Cooperation}
\label{sec:cooperate-federated}

\begin{figure}[h]
    \centering
    \includegraphics[width=0.8\linewidth,scale=1.00]{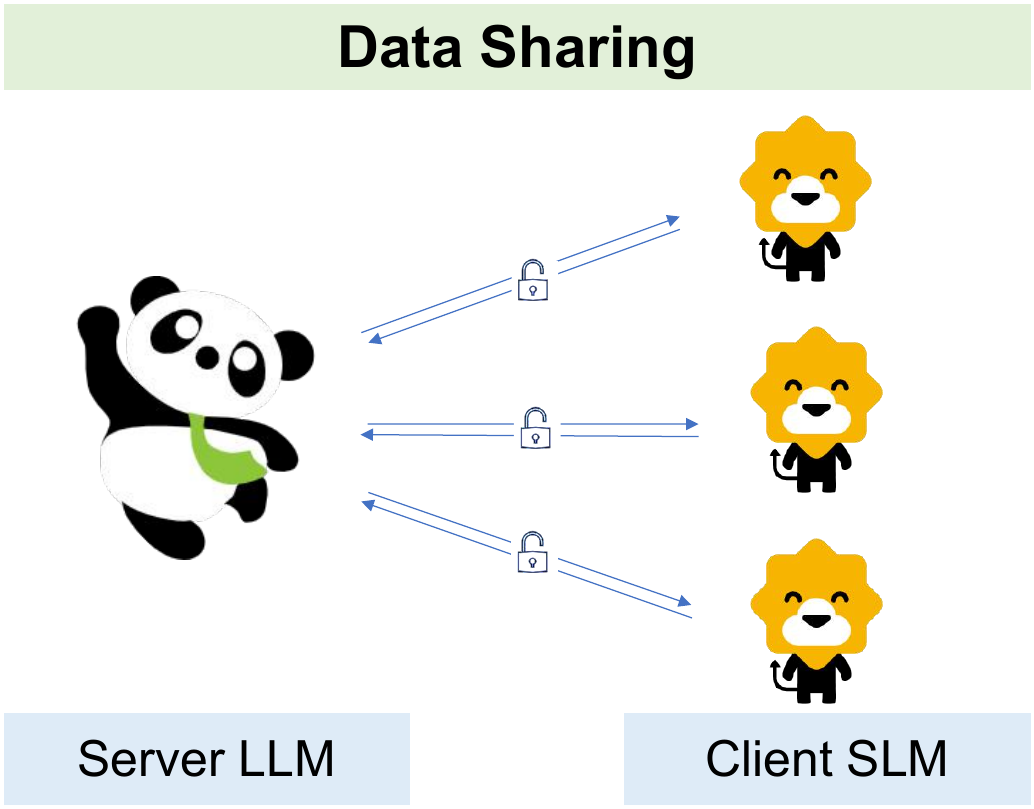}
    \caption{LLMs cooperate with client models in federated learning.}
    \label{fig:example-distribution}
\end{figure}

Large-scale language models have received increasing attention, but they face many challenges in real-world applications. These challenges mainly stem from the scarcity of public domain data and the need to maintain privacy for private data. To address these issues, federated learning (FL) \citep{Li_2020} has emerged as a very promising technology because it can protect private data while allowing the sharing of public models. Federated learning can be mainly divided into two parts: federated training and federated prompt engineering \citep{chen2023federated}. Existing collaborative studies are distributed across two aspects mentioned above.

\subsubsection{Federated Training}
\label{sec:cooperation-federate-training}
The knowledge of the LLMs on the server can be transferred to the small language models on the client, while the unique knowledge of the client can enrich the distribution of the LLMs. Consequently, the above methods greatly expand the pre-training corpora for both LLMs and small models \citep{Fan2024FedMKTFM, Ye2024OpenFedLLMTL}. To bridge the gap in the simultaneous mutual enhancement of both the servers' LLM and clients' SLMs, \citet{Fan2024FedMKTFM} develop FedMKT, a parameter-efficient federated mutual knowledge transfer framework for large and small language models. \citet{Ye2024OpenFedLLMTL} build a research-friendly framework, called OpenFedLLM, where multiple data owners collaboratively train a shared model without transmitting raw data. Additionally, to minimize resource consumption of parameters exchange, FedCyBGD \citep{Wang2024SaveIA} design a compression scheme  to further decrease the model download.

\subsubsection{Federated Prompt Engineering} 
\label{sec:cooperation-federate-prompt}
The goal of large model collaboration is to protect user privacy while efficiently executing commands. Therefore, using local small models to protect user privacy and cloud large models to execute user commands is a feasible approach \citep{Zhang2024CoGenesisAF, Li2024FederatedDK,guo2022promptfl}. For example, \citet{Zhang2024CoGenesisAF} deploy the small model locally to convert user instructions into general instructions. The large model in the cloud executes these general instructions and then returns the results to the local device. Because of discrepancies between LLMs' generated data and clients' domain-specific data, the exsiting methods cannot yield substantial improvements in the domain-specific tasks. Therefore, it introduces FDKT framework, which enables LLMs to augment data based on domain-specific few-shot demonstrations \citep{Li2024FederatedDK}. Furthermore, FL may cost excessive training time for convergence and produce inaccurate models. PromptFL \citep{guo2022promptfl} introduces prompt learner which updates the prompts instead of the whole model. Therefore, both the local training and the global aggregation can be significantly accelerated.



In addition to the aspects discussed above, we want to emphasize that cooperation among LLMs is a very broad research area. Topics such as RLHF and the use of agents also fall under LLM cooperation. As these topics have been extensively reviewed \cite{kaufmann2023survey, xi2023rise, sun2024llm, guo2024large}, we do not cover them in the current version of our paper but will address them in future work.

\section{Challenges and Future Directions}
\label{sec:future}

\subsection{Flexible LLMs Merging Methods}
Current LLM merging methods are typically restricted to models with same architecture and compatible parameters. However, most open-source LLMs are heterogeneous, rendering current merging methods ineffective.

Interestingly, \citet{xu2024bridging} demonstrate that the token embeddings of heterogeneous LLMs can be projected into a common space by using overlapping tokens as a bridge. However, this method cannot be successfully adapted to other parameters of LLMs, such as self-attention layers and feed-forward layers, due to the lack of aligned neurons and the complexity of parameter distributions. We believe that explore the highly correlated neurons \cite{singh2020model, pena2023re, ainsworth2023git, jordan2023repair, zipit} in diverse LLMs hold significant potential and interest. Such advancements could revolutionize model fusion techniques, enhancing their flexibility and practicality.

\subsection{Balanced Speed and Performance for LLM Ensemble}
As we discussed in \S \ref{sec:ensemble_discussion}, different LLM ensemble methods have their own strengths and weaknesses. Achieving a balance between speed and performance can be challenging.

LLM ensemble methods employed \textsc{Before} inference typically select a appropriate LLM for current example according to the pre-trained routers \cite{shnitzer2023large}. While these methods improve decoding speed, they are coarse-grained and do not fully harness the potential of ensemble learning. In contrast, methods used \textsc{During} inference operate at a finer granularity, achieving ensemble integration at the token level  \cite{xu2024bridging, huang2024enabling, yu2024breaking}. Such methods can alleviate errors at each decoding step via ensemble, relieving exploration bias \cite{ranzato2015sequence, xu2020rethinking} and holding greater promise for enhancing performance. However, these methods suffer from the slower inference speed. Effectively combining these two strategies may strike a balance between speed and performance improvements, benefiting practical deployments.

\subsection{Broader Applications via Cooperation}
This work primarily discusses several objectives achievable through cooperation among LLMs, including: efficient computation, knowledge transferring, compensatory cooperation and federated cooperation. Given the remarkable emergent capabilities of LLMs, we believe that a wide range of applications can be realized through flexible and judicious cooperation between different LLMs. For example, exploring cross-domain applications, where LLMs can combine their expertise in various fields will unlock new possibilities. Additionally, human-centered collaboration is also a promising direction \cite{ma2024model}.



\section{Conclusion}
This work presents a survey of collaboration strategies for LLMs, categorized into three aspects: \textit{merging}, \textit{ensemble}, and \textit{cooperation}. For each aspect, we provide a detailed classification and an in-depth review of advanced approaches. We believe that the collaboration of LLMs will play an increasingly important role in future research and hope that this paper offers valuable insights into future research directions.

\newpage
\section*{Contributions}
\ 

\textbf{Jinliang Lu} designed the overall architecture of this paper and was primarily responsible for \S \ref{sec:introduction}, \S \ref{sec:background}, \S \ref{sec:ensemble} and \S \ref{sec:cooperate-transferring}.

\textbf{Ziliang Pang} and \textbf{Yaochen Zhu} are responsible for the architecture of \S \ref{sec:merging}, Merging. Specifically, \textbf{Ziliang Pang} reviewed the literature in \S \ref{sec:merging-average}, \S \ref{sec:cooperate-accelerate}, and \textbf{Yaochen Zhu} reviewed the literature in \S \ref{sec:merging-multitask}.

\textbf{Min Xiao} is responsible for the architecture of \S \ref{sec:cooperation} Cooperation and reviewed the literature in \S \ref{sec:cooperate-compensatory} and \S \ref{sec:cooperate-federated}, primarily categorizing the corresponding studies based on their objectives.

\textbf{Jiajun Zhang} and \textbf{Rui Xia} led the project, designed and optimized the overall architecture of the survey, responsible for reviewing the entire work.


If you have suggestions or questions about this survey, please contact with us. We are very happy to hear from you.
\ 
\newpage
\bibliography{custom}

\appendix



\end{document}